\documentclass{article}

\PassOptionsToPackage{}{natbib}

    \usepackage[final]{neurips_2020}

\usepackage[utf8]{inputenc} 
\usepackage[T1]{fontenc}    

\usepackage[pagebackref=true,breaklinks=true,colorlinks=true,bookmarks=false,urlcolor=black, allcolors=mydarkblue]{hyperref}
\usepackage{url}            
\usepackage{booktabs}       
\usepackage{amsfonts}       
\usepackage{nicefrac}       
\usepackage{microtype}      

\usepackage{mathtools}  
\usepackage{caption}
\usepackage{subcaption}  
\usepackage{listings}   
\usepackage{booktabs}   
\usepackage{upgreek}
\usepackage[para]{footmisc}

\DeclareMathOperator*{\argmin}{arg\,min}
\DeclareMathOperator{\diag}{diag}
\DeclareMathOperator{\BatchNorm}{BatchNorm}

\newcommand\mat[1]{\mathbf{#1}}
\newcommand\mati[2]{\mat{#1}^{(#2)}}
\newcommand\matq[1]{\mat{\widehat{#1}}}
\newcommand\matsq[1]{\mat{\widetilde{#1}}}
\renewcommand\vec[1]{\mathbf{#1}}
\newcommand\veci[2]{\vec{#1}^{(#2)}}
\newcommand\vecq[1]{\vec{\widehat{#1}}}
\newcommand{\clamp}{\operatornamewithlimits{clamp}}
\newcommand\dw[0]{\Delta \! \vec{w}}
\newcommand\dW[0]{\mathrm{\Delta} \mat{W}}
\newcommand\ct[1]{#1}

\newcommand\bnbeta{\boldsymbol{\upbeta}}
\newcommand\bngamma{\boldsymbol{\upgamma}}
\newcommand\bnmu{\boldsymbol{\upmu}}
\newcommand\bnsigma{\boldsymbol{\upsigma}}

\newcommand{\rn}[1]{\text{ResNet{#1}}}
\newcommand{\mnv}[1]{\text{MobileNetV{#1}}}

\newcommand\func[2]{\mathnormal{#1}\left(#2\right)}
\newcommand\funcb[1]{\mathnormal{#1}}
\newcommand\tloss[1]{\mathcal{L}\left(#1\right)}
\newcommand\eop[1]{\mathop{\mathbb{E}}\left[#1\right]}

\DeclarePairedDelimiter\round{\lfloor}{\rceil}

\DeclarePairedDelimiter\floor{\lfloor}{\rfloor}
\DeclarePairedDelimiter\norm{\lVert}{\rVert}

\renewcommand{\l}{\left}
\renewcommand{\r}{\right}

\usepackage{changepage}

\makeatletter
\newcommand{\printfnsymbol}[1]{%
  \textsuperscript{\@fnsymbol{#1}}%
}
\makeatother

\makeindex

\usepackage{version}

\title{A White Paper on Neural Network Quantization}

\author{%
  Markus Nagel\thanks{Equal contribution.} \\ 
  Qualcomm AI Research\thanks{Qualcomm AI Research is an initiative of Qualcomm Technologies, Inc.}\\
  \texttt{markusn@qti.qualcomm.com}
  \And
  Marios Fournarakis\footnotemark[1] \\ 
  Qualcomm AI Research\footnotemark[2]\\
  \texttt{mfournar@qti.qualcomm.com} \AND
  Rana Ali Amjad \\ 
  Qualcomm AI Research\footnotemark[2]\\
  \texttt{ramjad@qti.qualcomm.com}
  \And
  Yelysei Bondarenko \\ 
  Qualcomm AI Research\footnotemark[2]\\
  \texttt{ybodaren@qti.qualcomm.com}
  \AND
  Mart van Baalen \\ 
  Qualcomm AI Research\footnotemark[2]\\
  \texttt{mart@qti.qualcomm.com}
  \And
  Tijmen Blankevoort \\ 
  Qualcomm AI Research\footnotemark[2]\\
  \texttt{tijmen@qti.qualcomm.com}
 }

\begin{document}

\maketitle

\begin{abstract}

While neural networks have advanced the frontiers in many applications, they often come at a high computational cost. Reducing the power and latency of neural network inference is key if we want to integrate modern networks into edge devices with strict power and compute requirements. Neural network quantization is one of the most effective ways of achieving these savings but the additional noise it induces can lead to accuracy degradation.

In this white paper, we introduce state-of-the-art algorithms for mitigating the impact of quantization noise on the network's performance while maintaining low-bit weights and activations. We start with a hardware motivated introduction to quantization and then consider two main classes of algorithms: Post-Training Quantization (PTQ) and Quantization-Aware-Training (QAT). PTQ requires no re-training or labelled data and is thus a lightweight push-button approach to quantization. In most cases, PTQ is sufficient for achieving  8-bit quantization with close to floating-point accuracy. QAT requires fine-tuning and access to labeled training data but enables lower bit quantization with competitive results. For both solutions, we provide tested pipelines based on existing literature and extensive experimentation that lead to state-of-the-art performance for common deep learning models and tasks.
\end{abstract}


\section{Introduction}
With the rise in popularity of deep learning as a general-purpose tool to inject intelligence into electronic devices, the necessity for small, low-latency and energy efficient neural networks solutions has increased. Today neural networks can be found in many electronic devices and services, from smartphones, smart glasses and home appliances, to drones, robots and self-driving cars. These devices are typically subject to strict time restrictions on the execution of neural networks or stringent power requirements for long-duration performance.

One of the most impactful ways to decrease the computational time and energy consumption of neural networks is quantization. In neural network quantization, the weights and activation tensors are stored in lower bit precision than the 16 or 32-bit precision they are usually trained in. When moving from 32 to 8 bits, the memory overhead of storing tensors decreases by a factor of 4 while the computational cost for matrix multiplication reduces quadratically by a factor of 16. Neural networks have been shown to be robust to quantization, meaning they can be quantized to lower bit-widths with a relatively small impact on the network's accuracy. Besides, neural network quantization can often be applied along with other common  methods for neural network optimization, such as neural architecture search, compression and pruning. It is an essential step in the model efficiency pipeline for any practical use-case of deep learning.
However, neural network quantization is not free. Low bit-width quantization introduces noise to the network that can lead to a drop in accuracy. While some networks are robust to this  noise, other networks require extra work to exploit the benefits of quantization. 

In this white paper, we introduce the state-of-the-art in neural network quantization. We start with an introduction to quantization and discuss  hardware and practical considerations.
We then consider two different regimes of quantizing neural networks: Post-Training Quantization (PTQ) and Quantization-Aware Training (QAT). PTQ methods, discussed in section~\ref{sec:ptq}, take a trained network and quantize it with little or no data, requires minimal hyperparameter tuning and no end-to-end training. This makes them a push-button approach to quantizing neural networks with low engineering effort and  computational cost. 
In contrast, QAT, discussed in section~\ref{sec:QAT}, relies on retraining the neural networks with simulated quantization in the training pipeline. While this requires more effort in training and potentially hyperparameter tuning, it generally further closes the gap to the full-precision accuracy compared to PTQ for low-bit quantization.
For both regimes, we introduce standard pipelines based on existing literature and extensive experimentation that lead to state-of-the-art performance for common computer vision and natural language processing models. We also propose a debugging workflow to identify and address common issues when quantizing a new model.

\section{Quantization fundamentals}
\label{sec:quant_intro}
In this section, we introduce the basic principles of neural network quantization and of fixed-point accelerators on which quantized networks run on. We start with a hardware motivation and then introduce standard quantization schemes and their properties. Later we discuss practical considerations related to layers commonly found in modern neural networks and their implications for fixed-point accelerators. 

\subsection{Hardware background}
\label{sec:hardware_background}
\begin{figure}[h]
\centering
\includegraphics[width=0.7\textwidth]{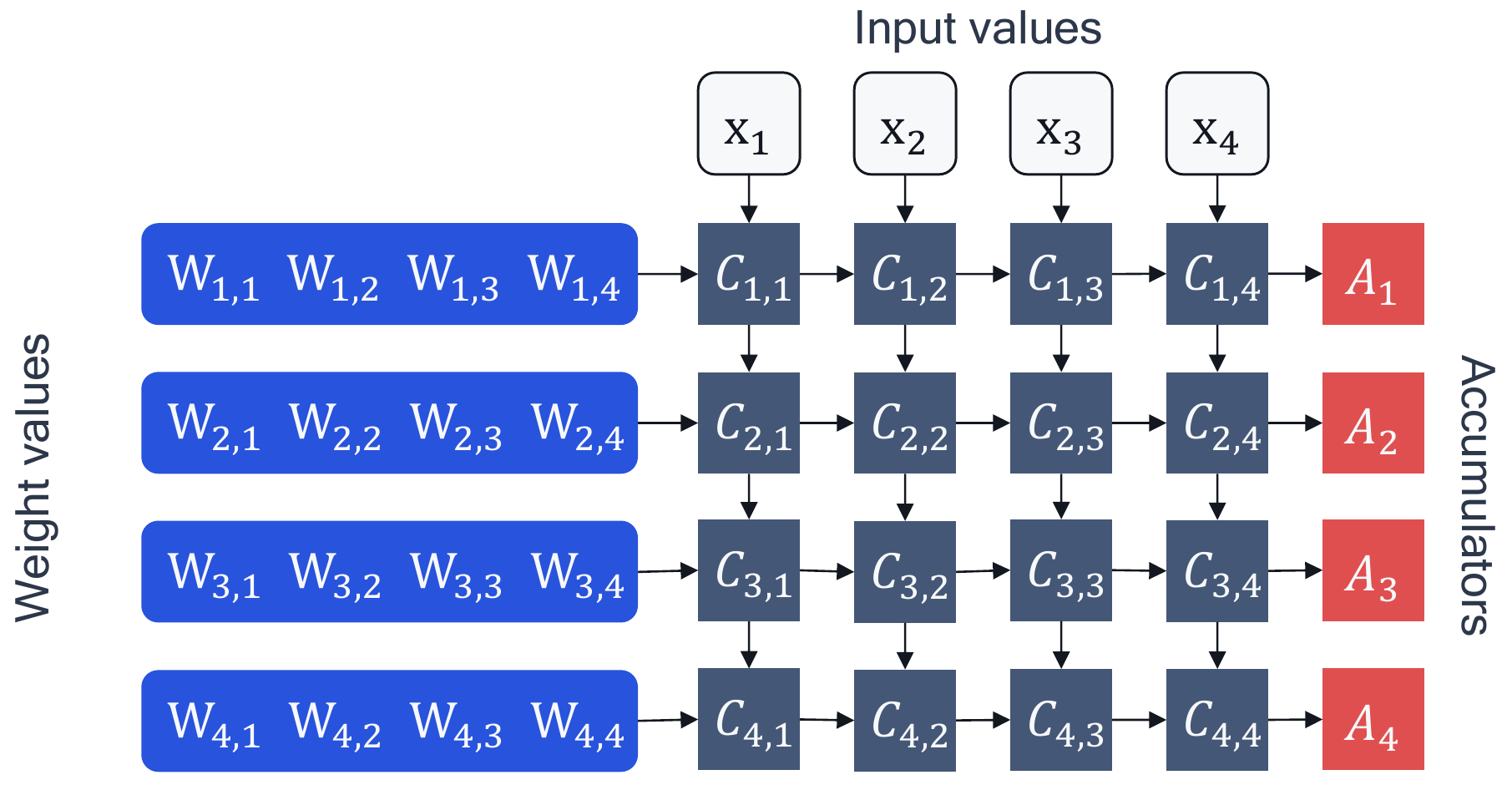}
\caption{A schematic overview of matrix-multiply logic in neural network accelerator hardware.}
\label{fig:nn_accelerator}
\end{figure}

Before diving into the technical details, we first explore the hardware background of quantization and how it enables efficient inference on device. Figure~\ref{fig:nn_accelerator} provides a schematic overview of how a matrix-vector multiplication, $\mat{y} = \mat{W} \vec{x} + \vec{b}$, is calculated in a neural network (NN) accelerator. This is the building block of larger matrix-matrix multiplications and convolutions found in neural networks. Such hardware blocks aim at improving the efficiency of NN inference by performing as many calculations as possible in parallel.
The two fundamental components of this NN accelerator are the \textit{processing elements} $\ct{C}_{n,m}$ and the \textit{accumulators} $\ct{A}_n$. Our toy example in figure \ref{fig:nn_accelerator} has 16 processing elements arranged in a square grid and $4$ accumulators. The calculation starts by loading the accumulators with the bias value $\vec{b}_n$.  We then load the weight values $\mat{W}_{n,m}$ and the input values $\vec{x}_m$ into the array and compute their product in the respective processing elements $\ct{C}_{n,m}=\mat{W}_{n,m} \,  \vec{x}_m$ in a single cycle. Their results are then added in the accumulators:
\begin{equation}
    \label{eq:accumulation}
    \ct{A}_n =\vec{b}_n + \sum_{m}{\ct{C}_{n,m}}
\end{equation}
The above operation is also referred to as \textit{Multiply-Accumulate} (MAC). This step is repeated many times for larger matrix-vector multiplications. Once all cycles are completed, the values in the accumulators are then moved back to memory to be used in the next neural network layer. 
Neural networks are commonly trained using FP32 weights and activations. If we were to perform inference in FP32, the processing elements and the accumulator would have to support floating-point logic, and we would need to transfer the 32-bit data from memory to the processing units.
MAC operations and data transfer consume the bulk of the energy spent during neural network inference. Hence, significant benefits can be achieved by using a lower bit fixed-point or \textit{quantized} representation for these quantities. Low-bit fixed-point representations, such as INT8, not only reduce the amount data transfer but also the size and energy consumption of the MAC operation \citep{horowitz}. This is because the cost of digital arithmetic typically scales linearly to quadratically with the number of bits used and because fixed-point addition is more efficient than its floating-point counterpart \citep{horowitz}.

To move from floating-point to the efficient fixed-point operations, we need a scheme for converting floating-point vectors to integers. A floating-point vector $\vec{x}$ can be expressed approximately as a  scalar multiplied by a vector of integer values:
\begin{equation}
\label{eq:simple_quant_example}
\widehat{\vec{x}}  = \ct{s}_{\vec{x}} \cdot \vec{x}_{\text{int}} \approx  \vec{x}
\end{equation}
where $\ct{s}_{\vec{x}}$ is a floating-point \textit{scale factor} and  $\vec{x}_{\text{int}}$ is an integer vector, e.g., INT8. We denote this \textit{quantized} version of the vector as $\widehat{\vec{x}}$. By quantizing the weights and activations we can write the quantized version of the accumulation equation:
\begin{align}
  \hat{\ct{A}}_n& = \widehat{\vec{b}}_n +  \sum_m{\widehat{\mat{W}}_{n,m} \, \widehat{\vec{x}}_m} \nonumber \\
  & =  \widehat{\vec{b}}_n +  \sum_m{ \l(\ct{s}_{\vec{w}} \mat{W}^{\text{int}}_{n,m} \r)\l(\ct{s}_{\vec{x}}\vec{x}^{\text{int}}_m \r)} \nonumber \\
  & =  \widehat{\vec{b}}_n +   \ct{s}_{\vec{w}} \ct{s}_{\vec{x}} \sum_{m}{\mat{W}^{\text{int}}_{n,m} \,\vec{x}^{\text{int}}_m}  \label{eq:quantized_accumulation} 
\end{align}
Note that we used a separate scale factor for weights, $ \ct{s}_{\vec{w}}$, and activations, $\ct{s}_{\vec{x}}$. This provides flexibility and reduces the quantization error (more in section \ref{sec:quant_scheme}). Since each scale factor is applied to the whole tensor, this scheme allows us to factor the scale factors out of the summation in equation \eqref{eq:quantized_accumulation} and perform MAC operations in fixed-point format. We intentionally ignore bias quantization for now, because the bias is normally stored in higher bit-width (32-bits) and its scale factor depends on that of the weights and activations \citep{jacob2018cvpr}.

\begin{figure}[h]
\centering
\includegraphics[width=0.9\textwidth]{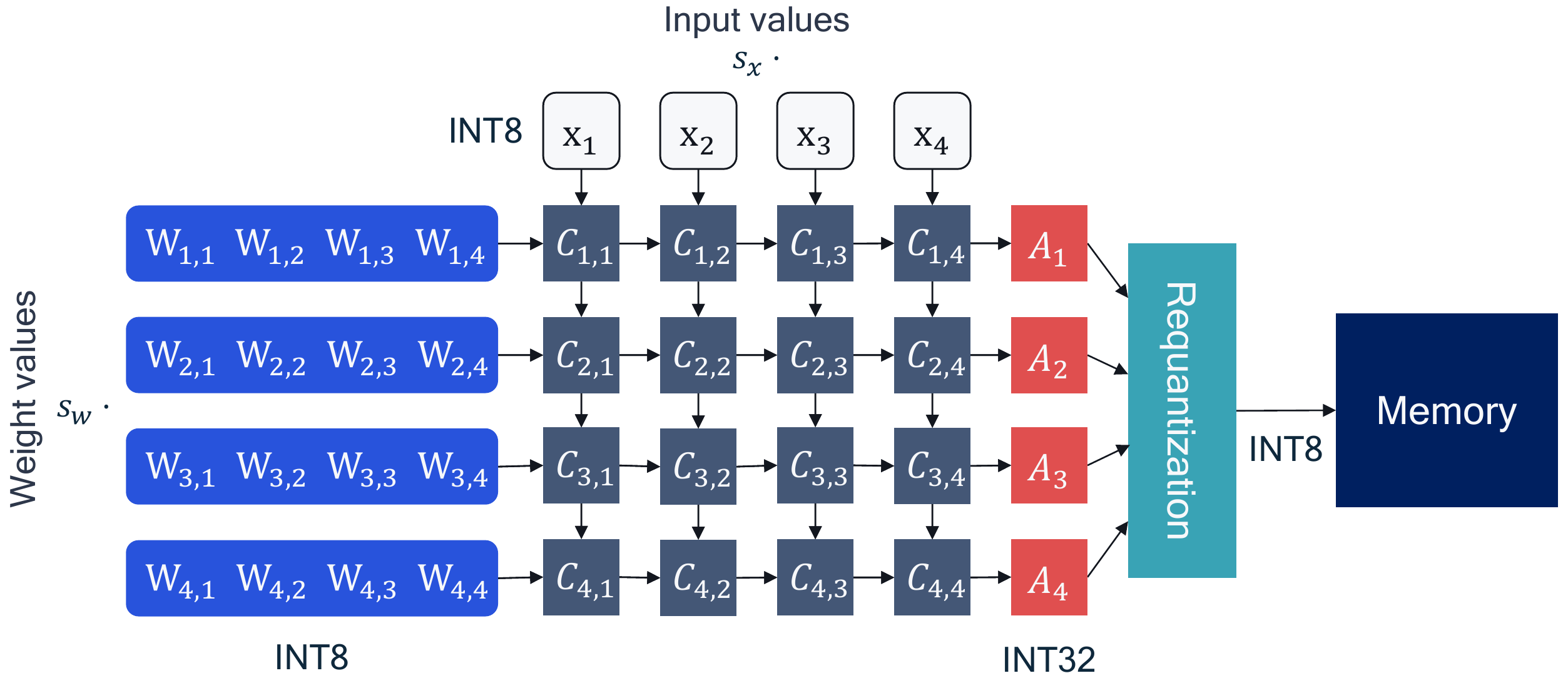}
\caption{A schematic of matrix-multiply logic in an neural network accelerator for quantized inference.}
\label{fig:quanthardware}
\end{figure}

Figure \ref{fig:quanthardware} shows how the neural network accelerator changes when we introduce quantization. In our example, we use INT8 arithmetic, but this could be any quantization format for the sake of this discussion. It is important to maintain a higher bit-width for the accumulators, typical 32-bits wide. Otherwise, we risk incurring loss due to overflow as more products are accumulated during the computation.

The activations stored in the 32-bit accumulators need to be written to memory before they can be used by the next layer. To reduce data transfer and the complexity of the next layer's operations, these activations are quantized back to INT8. This requires a \textit{requantization} step which is  shown in figure \ref{fig:quanthardware}.
 
\subsection{Uniform affine quantization}
\label{sec:quant_scheme}
In this section we define the quantization scheme that we will use in this paper. This scheme is called \textit{uniform quantization} and it is the most commonly used quantization scheme because it permits efficient implementation of fixed-point arithmetic.

\textit{Uniform affine quantization}, also known as \textit{asymmetric quantization}, is defined by three quantization  parameters: the \textit{scale factor} $ 
\ct{s}$, the \textit{zero-point} $\ct{z}$ and the \textit{bit-width} $\ct{b}$. The scale factor and the zero-point are used to to map a floating point value to the integer grid, whose size depends on the bit-width. The scale factor is commonly represented as a floating-point number and specifies the \textit{step-size} of the quantizer. The zero-point is an integer that ensures that real zero is quantized without error. This is important to ensure that common operations like zero padding or ReLU do not induce quantization error.

Once the three quantization parameters are defined we can proceed with the quantization operation. Starting from a real-valued vector $\vec{x}$ we first map it to the \textit{unsigned} integer grid $\{0,\dots,  2^\ct{b}-1\}$:
\begin{equation}
\label{eq:quant_operation}
    \vec{x}_{\text{int}} =\clamp{\left(\round*{\frac{\vec{x}}{\ct{s}}} + \ct{z} ;0, 2^\ct{b}-1 \right)},
\end{equation}
where $\round*{\cdot}$ is the round-to-nearest operator and clamping is defined as:
\begin{equation}
\label{eq:clamping}
\clamp\left(\ct{x}; \ct{a}, \ct{c}\right) =
    \begin{cases}
            \ct{a}, & \quad \ct{x} <  \ct{a}, \\
            \ct{x}, &  \quad  \ct{a}\leq x  \leq \ct{c},\\
             \ct{c}, & \quad x> \ct{c}.
    \end{cases}
\end{equation}
To approximate the real-valued input $\vec{x}$ we perfrom a  \textit{de-quantization} step:
\begin{equation}
    \label{eq:de_quantization}
        \vec{x}\approx \widehat{\vec{x}}  = \ct{s}\l(\vec{x}_{\text{int}}-\ct{z}\r)
\end{equation}
Combining the two steps above we can provide a general definition for the \textit{quantization function}, $q(\cdot)$, as:
\begin{equation}
    \label{eq:quant_function}
    \widehat{\vec{x}}= q(\vec{x};\ct{s},\ct{z}, \ct{b}) = \ct{s}\left[\clamp{\left(\round*{\frac{\vec{x}}{\ct{s}}}+\ct{z};0 ,2^\ct{b}-1  \right)}-\ct{z}\right], 
\end{equation}
Through the de-quantization step, we can also define the quantization grid limits $(\ct{q}_{\text{min}}, \ct{q}_{\text{max}})$ where $\ct{q}_{\text{min}} = -\ct{s}\ct{z}$ and  $\ct{q}_{\text{max}}=\ct{s}(2^\ct{b}-1-\ct{z})$. Any values of $\vec{x}$ that lie outside this range will be clipped to its limits, incurring a \textit{clipping error}. If we want to reduce the clipping error we can expand the quantization range by increasing the scale factor. However, increasing the scale factor leads to increased \textit{rounding error} as the rounding error lies in the range $\l[-\frac{1}{2}\ct{s}, \frac{1}{2}\ct{s}\r]$. In section \ref{sec:range_setting}, we explore in more detail how to choose the quantization parameters to achieve the right trade-off between clipping and rounding errors.

\subsubsection{Symmetric uniform quantization}
\label{sec:symmetric_quantizer}
\begin{figure}[h]
\centering
\includegraphics[width=0.85\textwidth]{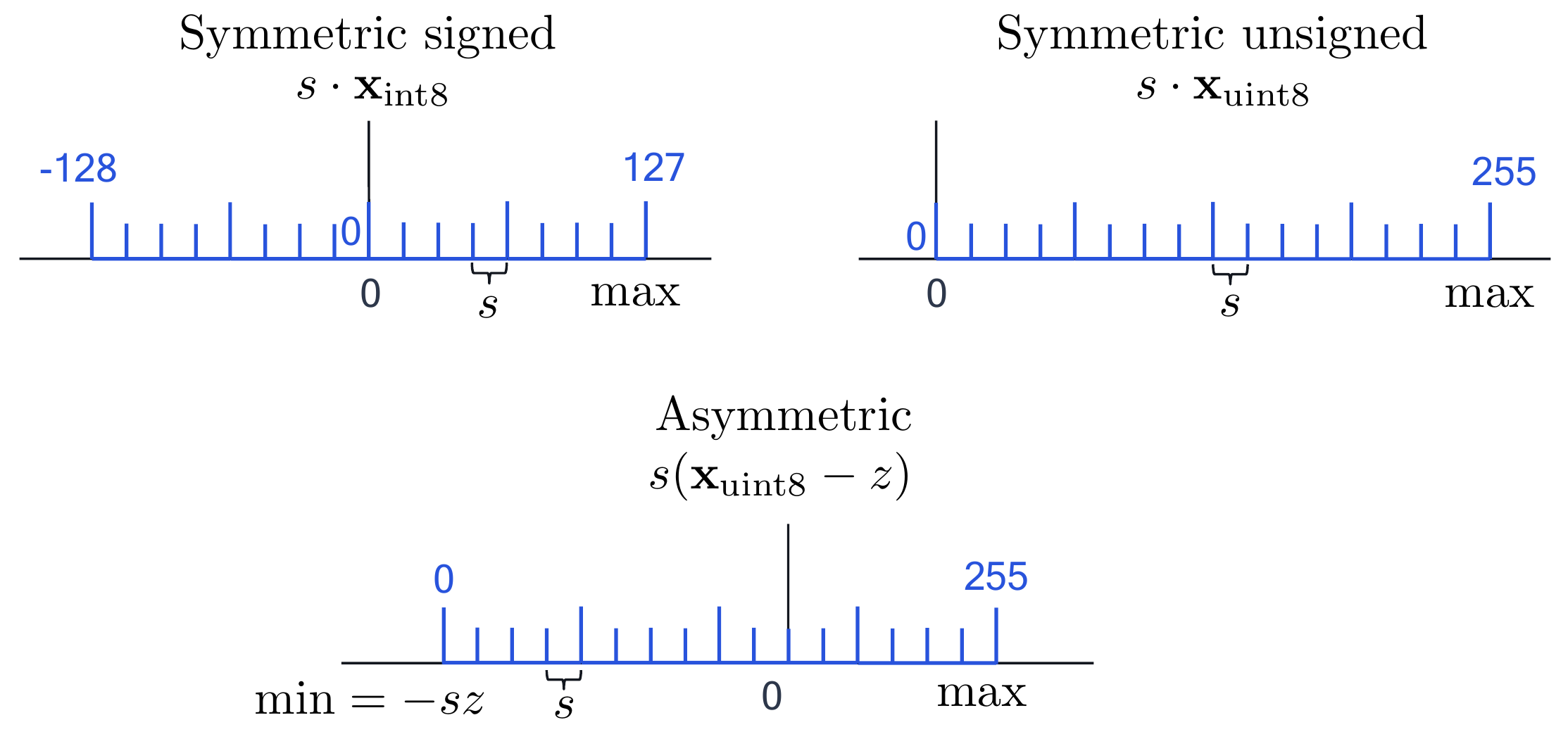}
\caption{A visual explanation of the different uniform quantization grids for a bit-width of 8. $\ct{s}$ is the scaling factor, $\ct{z}$ the zero-point. The floating-point grid is in black, the integer quantized grid in blue.}
\label{fig:quantization_schemes}
\end{figure}

Symmetric quantization is a simplified version of the general asymmetric case. The symmetric quantizer restricts the zero-point to 0. This reduces the computational overhead of dealing with zero-point offset during the accumulation operation in equation \eqref{eq:quantized_accumulation}. But the lack of offset restricts the mapping between integer and floating-point domain. As a result, the choice of signed or unsigned integer grid matters:
\begin{subequations}
\label{eq:symmetric_quantization}
\begin{align}
     \widehat{\vec{x}} &= \ct{s} \, \vec{x}_{\text{int}} \\
     \vec{x}_{\text{int}}  &= \clamp\l(\round*{\frac{\vec{x}}{\ct{s}}};0, 2^\ct{b}-1 \r) & \text{for unsigned integers}\\
     \vec{x}_{\text{int}}  &= \clamp\l(\round*{\frac{\vec{x}}{\ct{s}}};-2^{\ct{b}-1} , 2^{\ct{b}-1}-1\r)& \text{for signed integers} 
\end{align}
\end{subequations}
Unsigned symmetric quantization is well suited for one-tailed distributions, such as ReLU activations (see figure \ref{fig:quantization_schemes}). On the other hand, signed symmetric quantization can be chosen for distributions that are roughly symmetric about zero.

\subsubsection{Power-of-two quantizer}
\label{sec:power_of_two_quantizer}
Power-of-two quantization is a special case of symmetric quantization, in which the scale factor is restricted to a power-of-two, $\ct{s}=2^{-\ct{k}}$. This choice can bring hardware efficiencies because scaling with $s$ corresponds to simple bit-shifting.  However, the restricted expressiveness of the scale factor can complicate the trade-off between rounding and clipping error.

\subsubsection{Quantization granularity}
\label{sec:quant_granularity}
So far, we have defined a single set of quantization parameters (quantizer) per tensor, one for the weights and one for activations, as seen in equation \eqref{eq:quantized_accumulation}. This is called \textit{per-tensor quantization}. We can also define a separate quantizer for individual segments of a tensor (e.g., output channels of a weight tensor), thus increasing the \textit{quantization granularity}. In neural network quantization, per-tensor quantization is the  the most common choice of granularity due to its simpler hardware implementation: all accumulators in equation \eqref{eq:quantized_accumulation} use the same scale factor, $\ct{s}_{\vec{w}}\ct{s}_{\vec{x}}$. However, we could use finer granularity to further improve performance. For example, for weight tensors, we can specify a different quantizer per output channel. This is known as \textit{per-channel} quantization and its implications are discussed in more detailed in section \ref{sec:per_channel_quantization}. 

Other works go beyond per-channel quantization parameters and apply separate quantizers per group of weights or activations \citep{MSFT,theBitGoesDown,dsConv}.  Increasing the granularity of the groups generally improves accuracy at the cost of some extra overhead. The overhead is associated with accumulators handling  sums of values with varying scale factors. Most existing fixed-point accelerators do not currently support such logic and for this reason, we will not consider them in this work.  However, as research in this area grows, more hardware support for these methods can be expected in the future.

\subsection{Quantization simulation}
\label{sec:quantization_simulation}
To test how well a neural network would run on a quantized device, we often simulate the quantized behavior on the same general purpose hardware we use for training neural networks. This is called \textit{quantization simulation}. We aim to  approximate fixed-point operations using floating-point hardware. 
Such simulations are significantly easier to implement compared to running experiments on actual quantized hardware or using quantized kernels. They allow the user to efficiently test various quantization options and it enables GPU acceleration for quantization-aware training as described in section \ref{sec:QAT}. In this section, we first explain the fundamentals of this simulation process and then discuss techniques that help to reduce the difference between the simulated and the actual on-device performance.
\begin{figure}
     \centering
     \begin{subfigure}[b]{0.46\textwidth}
         \centering
         \includegraphics[width=0.75\textwidth]{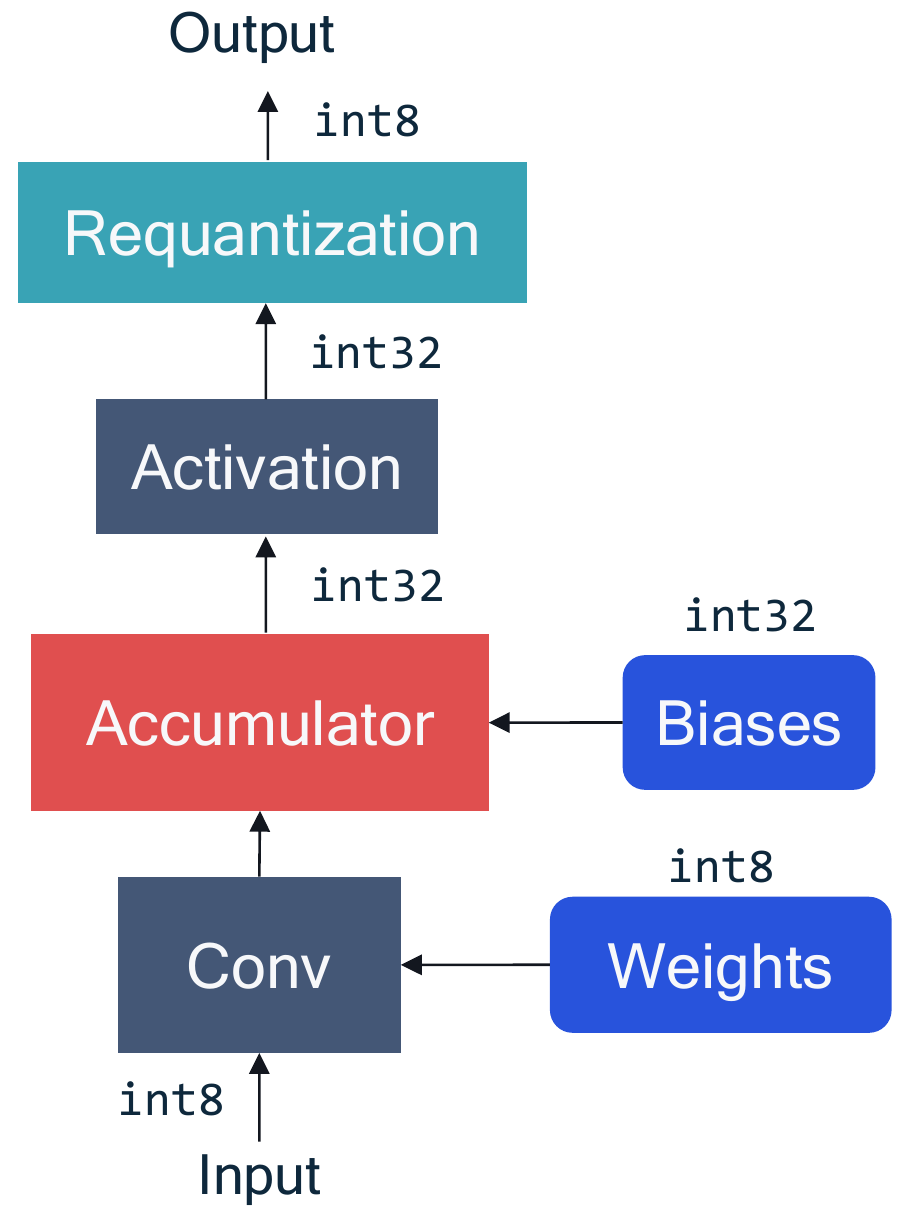}
         \caption{Diagram for quantized on-device inference with fixed-point operations.}
         \label{fig:on-device_quant_inference}
     \end{subfigure}
     \hfill
     \begin{subfigure}[b]{0.44\textwidth}
         \centering
         \includegraphics[width=\textwidth]{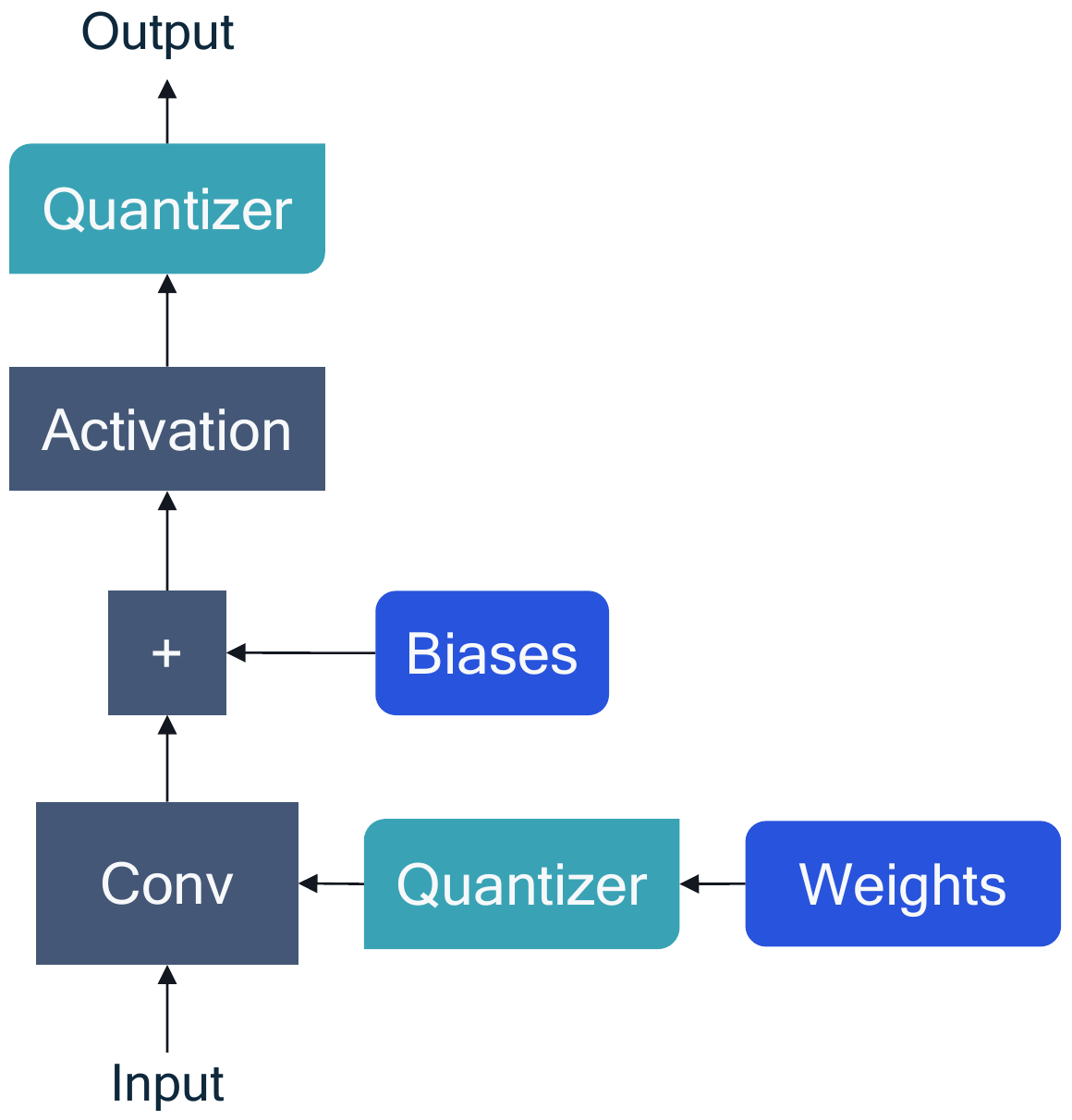}
         \caption{Simulated quantization using floating-point operations.}
         \label{fig:simulated quantization}
     \end{subfigure}
        \caption{Schematic overview of quantized forward pass for convolutional layer: a) Compute graph of actual on-device quantized inference. b) Simulation of quantized inference for general-purpose floating-point hardware.}
        \label{fig:simulation}
\end{figure}

Previously, we saw how matrix-vector multiplication is calculated in dedicated fixed-point hardware. In figure~\ref{fig:on-device_quant_inference}, we generalize this process for a convolutional layer, but we also include an activation function to make it more realistic. During on-device inference, all the inputs (biases, weight and input activations) to the hardware are in a fixed-point format. However, when we simulate quantization using common deep learning frameworks and general-purpose hardware these quantities are in floating-point. This is why we introduce quantizer blocks in the compute graph to induce quantization effects. 

Figure \ref{fig:simulated quantization} shows how the same convolutional layer is modelled in a deep-learning framework. Quantizer blocks are added in between the weights and the convolution to simulate weight quantization, and after the activation function to simulate activation quantization. The bias is often not quantized because it is stored in higher-precision. In section \ref{sec:quant_op_fusing}, we discuss in more detail when it is appropriate to position the quantizer after the non-linearity. The quantizer block implements the quantization function of equation \eqref{eq:quant_function} and each quantizer is defined by a set of quantization parameters (scale factor, zero-point, bit-width). Both the input and output of the quantizer are in floating-point format but the output lies on the quantization grid. 

\subsubsection{Batch normalization folding}
\label{sec:batch_norm_folding}
Batch normalization \citep{batchnorm} is a standard component of modern convolutional networks. Batch normalization normalizes the output of a linear layer before scaling and adding an offset (see equation~\ref{eq:batch_norm}). For on-device inference, these operations are folded into the previous or next linear layers in a step called \textit{batch normalization folding} \citep{krishnamoorthi,jacob2018cvpr}. This removes the batch normalization operations entirely from the network, as the calculations are absorbed into an adjacent linear layer. Besides reducing the computational overhead of the additional scaling and offset, this prevents extra data movement and the quantization of the layer's output.
More formally, during inference, batch normalization is defined as an affine map of the output $\vec{x}$:
\begin{equation}
\label{eq:batch_norm}
 \BatchNorm(\vec{x}) = \gamma\left(\frac{\vec{x} - \mu}{\sqrt{\sigma^2 + \epsilon}}\right) + \beta
\end{equation}
where $\mu$ and $\sigma$ are the mean and variance computed during training as exponential moving average over batch-statistics, and $\gamma$ and $\beta$ are learned affine hyper-parameters per-channel. If batch normalization is applied right after a linear layer $\vec{y} = \BatchNorm(\mat{W}\vec{x})$, we can rewrite the terms such that the batch normalization operation is fused with the linear layer itself. Assuming a weight matrix $\mat{W}\in \mathbb{R}^{\ct{n}\times\ct{m}}$ we apply batch normalization to each output $\vec{y}_k$ for $k=\{1,\dots, n\}$:
\begin{align}
        \vec{y}_k &= \BatchNorm(\mat{W}_{k,:} \, \vec{x}) \nonumber \\
        &= \bngamma_k \left(\frac{\mat{W}_{k,:} \,\vec{x} - \bnmu_k}{\sqrt{\bnsigma_k^2 + \epsilon}}\right) + \bnbeta_k \nonumber \\
 &= \frac{\bngamma_k \mat{W}_{k,:}}{\sqrt{\bnsigma_k^2 + \epsilon}} \vec{x} + \l( \bnbeta_k - \frac{\bngamma_k \bnmu_k}{\sqrt{\bnsigma_k^2 + \epsilon}}\r)  \nonumber \\
 &= \widetilde{\mat{W}}_{k,:}\, \vec{x} + \widetilde{\vec{b}}_k\label{eq:bn_folding}
\end{align}
where:
\begin{align}
    \widetilde{\mat{W}}_{k,:} & =  \frac{\bngamma_k \mat{W}_{k,:}}{\sqrt{\bnsigma_k^2 + \epsilon}} , \label{eq:folded_weights}\\
    \widetilde{\vec{b}}_k & = \bnbeta_k - \frac{\bngamma_k \bnmu_k}{\sqrt{\bnsigma_k^2 + \epsilon}} . \label{eq:folded_bias}
\end{align}


\subsubsection{Activation function fusing}
\label{sec:quant_op_fusing}

In our naive quantized accelerator introduced in section \ref{sec:hardware_background}, we saw that the requantization of activations happens after the matrix multiplication or convolutional output values are calculated. However, in practice, we often have a non-linearity directly following the linear operation. It would be  wasteful to write the linear layer's activations to memory, and then load them back into a compute core to apply a non-linearity. For this reason, many hardware solutions come with a hardware unit that applies the non-linearity before the requantization step. If this is the case, we only have to simulate requantization that happens after the non-linearity. For example, ReLU non-linearities are readily modelled by the requantization block, as you can just set the minimum representable value of that activation quantization to 0.

Other more complex activation functions, such as sigmoid or Swish \citep{swish},  require more dedicated support. If this support is not available, we need to add a quantization step before and after the non-linearity in the graph. This can have a big impact on the accuracy of quantized model. Although newer activations like Swish functions provide accuracy improvement in floating-point, these may vanish after quantization or may be less efficient to deploy on fixed-point hardware.

\subsubsection{Other layers and quantization}
\label{sec:other_layers_and_quantization}
There are many other types of layers being used in neural networks. How these are modeled depends greatly on the specific hardware implementation. Sometimes the mismatch between simulated quantization and on-target performance is down to layers not being properly quantized. Here, we provide some guidance on how to simulate quantization for a few commonly used layers:
\begin{description}
    \item[Max pooling] Activation quantization is not required because the input and output values are on the same quantization grid.
    \item[Average pooling] The average of integers is not  necessarily an integer. For this reason, a quantization step is required after average-pooling. However, we use the same quantizer for the inputs and outputs as the quantization range does not significantly change.
    \item[Element-wise addition] Despite its simple nature, this operation is difficult to simulate accurately. During addition, the quantization ranges of both inputs have to match exactly. If these ranges do not match, extra care is needed to make addition work as intended. There is no single accepted solution for this but adding a requantization step can simulate the added noise coarsely. Another approach is to optimize the network by tying the quantization grids of the inputs. This would prevent the requantization step but may require fine-tuning. 
    \item[Concatenation] The two branches that are being concatenated generally do not share the same quantization parameters. This means that their quantization grids may not overlap making a requantization step necessary. As with element-wise addition, it is possible to optimize your network to have shared quantization parameters for the branches being concatenated. 
\end{description}

\subsection {Practical considerations} 
\label{sec:practical considerations}
When quantizing neural networks with multiple layers, we are confronted with a large space of quantization choices including the quantization scheme, granularity, and bit-width. In this section, we explore some of the practical considerations that help reduce the search space.

Note that in this white paper we only consider \textit{homogeneous} bit-width. This means that the bit-width chosen for either weights or activations remains constant across all layers. Homogeneous bit-width is more universally supported by hardware but some recent works also explore the implementation of \textit{heterogeneous} bit-width or \textit{mixed-precision} \citep{bayesianbits,hawq,differentiablequantization}.

\subsubsection{Symmetric vs. asymmetric quantization}
\label{sec:symmetric_vs_asymmetric_quant}
For each weight and activation quantization, we have to choose a quantization scheme. On one hand, asymmetric quantization is more expressive because there is an extra offset parameter, but on the other hand there is a possible computational overhead. To see why this is the case, consider what happens when asymmetric weights, $\widehat{\mat{W}} = \ct{s}_{\vec{w}}(\mat{W}_{\text{int}} - \ct{z}
_{\vec{w}})$, are multiplied with asymmetric activations $\widehat{\mat{x}} = \ct{s}_{\vec{x}}(\mat{x}_{\text{int}} - \ct{z}
_{\vec{x}})$:
\begin{align}
    \widehat{\mat{W}} \widehat{\mat{x}} & = \ct{s}_{\vec{w}} (\mat{W}_{\text{int}} - \ct{z}
_{\vec{w}})  \ct{s}_{\vec{x}}(\mat{x}_{\text{int}} - \ct{z}
_{\vec{x}}) \nonumber \\
    & = \ct{s}_{\vec{w}}  \ct{s}_{\vec{x}} \mat{W}_{\text{int}} \mat{x}_{\text{int}}  - \textcolor{red}{\ct{s}_{\vec{w}} \ct{z}
_{\vec{w}}  \ct{s}_{\vec{x}} \mat{x}_{\text{int}}} - \textcolor{blue}{\ct{s}_{\vec{w}}  \ct{s}_{\vec{x}} \ct{z}
_{\vec{x}} \mat{W}_{\text{int}} + \ct{s}_{\vec{w}} \ct{z}
_{\vec{w}} \ct{s}_{\vec{x}} \ct{z}
_{\vec{x}}}. \label{eq:asymmetric_integer_matmult}
\end{align}
The first term is what we would have if both operations were in symmetric format. The third and fourth terms depend only on the scale, offset and weight values, which are known in advance. Thus these two terms can be pre-computed and added to the bias term of a layer at virtually no cost. The second term, however, depends on the input data $\mat{x}$. This means that for each batch of data we need to compute an additional term during inference. This can lead to significant overhead in both latency and power, as it is equivalent to adding an extra channel.

For this reason, it is a common approach to use \textit{asymmetric activation quantization} and \textit{symmetric weight quantization} that avoids the additional data-dependent term. 

\subsubsection{Per-tensor and per-channel quantization}
\label{sec:per_channel_quantization}
In section \ref{sec:quant_granularity}, we discussed different levels of quantization granularity. Per-tensor quantization of weights and activations has been standard for a while because it is supported by all fixed-point accelerators. However, per-channel quantization of the weights can improve accuracy, especially when the distribution of weights varies significantly from channel to channel. Looking back at the quantized MAC operation in equation \eqref{eq:quantized_accumulation}, we can see that per-channel weight quantization can be implemented in the accelerator by applying a separate per-channel weight scale factor without requiring rescaling. Per-channel quantization of activations is much harder to implement because we cannot factor the scale factor out of the summation and would, therefore, require rescaling the accumulator for each input channel. Whereas \textit{per-channel quantization} of weights is increasingly becoming common practice, not all commercial hardware supports it. Therefore, it is important to check if it is possible in your intended target device.


\section{Post-training quantization}
\label{sec:ptq}
Post-training quantization (PTQ) algorithms take a pre-trained FP32 network and convert it directly into a fixed-point network without the need for the original training pipeline. These methods can be data-free or may require a small calibration set, which is often readily available. Additionally, having almost no hyperparameter tuning makes them usable via a single API call as a black-box method to quantize a pretrained neural network in a computationally efficient manner. 
This frees the neural network designer from having to be an expert in quantization and thus allows for a much wider application of neural network quantization.

A fundamental step in the PTQ process is finding good quantization ranges for each quantizer. We briefly discussed in section~\ref{sec:quant_scheme} how the choice of quantization range affects the quantization error.
In this section, we start by discussing various common methods used in practice to find good quantization parameters. We then explore common issues observed during PTQ and introduce the most successful techniques to overcome them. Using these techniques we present a standard post-training quantization pipeline, which we find to work best in most common scenarios and, finally, we introduce a set of debugging steps to improve the performance of the quantized model.

\subsection{Quantization range setting}
\label{sec:range_setting}
Quantization range setting refers to the method of determining clipping thresholds of the quantization grid, $q_{\text{min}}$ and $q_{\text{max}}$ (see equation~\ref{eq:quant_function}). The key trade-off in range setting is between clipping and rounding error, described in section~\ref{sec:quant_scheme},  and their impact on the final task loss for each quantizer being configured. Each of the methods described here provides a different trade-off between the two quantities. These methods typically optimize local cost functions instead of the task loss. This is because in PTQ we aim for computationally fast methods without the need for end-to-end training.

Weights can usually be quantized without any need for calibration data. However, determining parameters for activation quantization often requires a few batches of calibration data.

\paragraph{\textbf{Min-max}} To cover the whole dynamic range of the tensor, we can define the quantization parameters as follows 
\begin{align}
    q_{\text{min}} &= \min \mat{V}, \\
    q_{\text{max}} &= \max \mat{V},
\end{align}
where $\mat{V}$ denotes the tensor to be quantized. This leads to no clipping error. 
However, this approach is sensitive to outliers as strong outliers may cause excessive rounding errors.

\paragraph{\textbf{Mean squared error (MSE)}}
One way to alleviate the issue of large outliers is to use MSE-based range setting. 
In this range setting method we find $q_\text{min}$ and $q_\text{max}$ that minimize the MSE between the original and the quantized tensor:
\begin{align}
    \argmin_{q_{\text{min}}, q_{\text{max}}} \quad \norm*{\mat{V} - \hat{\mat{V}}(q_{\text{min}}, q_{\text{max}})}^2_F, \label{eq:rangesetting_mse}
\end{align}
where $\hat{\mat{V}}(q_{\text{min}}, q_{\text{max}})$ denotes the quantized version of $\mat{V}$ and $\norm*{\cdot}_F$ is the Frobenius norm. 
The optimization problem is commonly solved using grid search, golden section method or analytical approximations with closed-form solution \citep{bannerposttraining}. Several variants of this range setting method exist in literature but they are all very similar in terms of objective function and optimization.

\begin{table*}[t]
    \centering
    \begin{tabular}{ l | r r r | r r r}
    \toprule
        Model (FP32 accuracy) & \multicolumn{3}{c|}{\rn{18} (69.68)} & \multicolumn{3}{c}{\mnv{2} (71.72)} \\ \midrule
        Bit-width        & \multicolumn{1}{c}{W8}    & \multicolumn{1}{c}{W6}      & \multicolumn{1}{c|}{W4}    & \multicolumn{1}{c}{W8} & \multicolumn{1}{c}{W6}    & \multicolumn{1}{c}{W4} \\\midrule
Min-Max                 & \textbf{69.57} &	63.90 &	0.12 &	\textbf{71.16} &	64.48 &	0.59 \\
MSE                     & 69.45 &	\textbf{64.64} &	\textbf{18.82} &	71.15 &	\textbf{65.43} &	\textbf{13.77} \\ \midrule
Min-Max (Per-channel)   &69.60 &	69.08 &	44.49 &	71.21 &	68.52 &	18.40 \\
MSE (Per-channel)       &\textbf{69.66} &	\textbf{69.24} &	\textbf{54.67} &	\textbf{71.46} &	\textbf{68.89} &	\textbf{27.17} \\
\bottomrule
    \end{tabular}
\caption{Ablation study for different methods of range setting of (symmetric uniform) weight quantizers while keeping the activations in FP32. Average ImageNet validation accuracy (\%) over 5 runs.}
 \label{tab:rangesetting_weights}
\end{table*}


\paragraph{\textbf{Cross entropy}}
For certain layers, all values in the tensor being quantized may not be equally important. One such scenario is the quantization of logits in the last layer of classification networks, in which it is important to preserve the order of the largest value after quantization. MSE may not be a suitable metric for this, as it weighs all the values in a tensor equally regardless of their order. For a larger number of classes, we usually have a large number of small or negative logits that are unimportant for prediction accuracy and few larger values that matter. In this case, MSE would incur a large quantization error to the few larger important logits while trying to reduce the quantization error of the more populous smaller logits.
In this specific case, it is beneficial to minimize the following cross-entropy loss function
\begin{align}
    \argmin_{q_{\text{min}}, q_{\text{max}}} \quad H\left(\psi\left(\vec{v}\right), \psi\left(\hat{\vec{v}}(q_{\text{min}},
    q_{\text{max}})\right)\right)
\end{align}
where $H(\cdot,\cdot)$ denotes the cross-entropy function, $\psi$ is the softmax function, and $\vec{v}$ is the logits vector.

\paragraph{\textbf{BN based range setting}}
Range setting for activation quantizers often requires some calibration data. 
If a layer has batch-normalized activations, the per-channel mean and standard deviation of the activations are equal to the learned batch normalization shift and scale parameters, respectively. 
These can then be used to find suitable parameters for activation quantizer as follows \citep{dfq}:
\begin{align}
 q_{\text{min}} &= \min{(\bnbeta - \alpha \bngamma)} \\
 q_{\text{max}} &= \max{( \bnbeta + \alpha \bngamma )}
\end{align}
where $\bnbeta$ and $\bngamma$ are vectors of per-channel learned shift and scale parameters, and $\alpha > 0$. \citet{dfq} uses $\alpha=6$ so that only large outliers are clipped.

\paragraph{\textbf{Comparison}}
In table \ref{tab:rangesetting_weights} we compare range setting methods for weight quantization. 
For high bit-widths, the MSE and min-max approaches are mostly on par. However, at lower bit-widths the MSE approach clearly outperforms the min-max.
In table \ref{tab:rangesetting_activations}, we present a similar comparison for activation quantization. We note that MSE combined with cross-entropy for the last layer, denoted as MSE + Xent, outperforms other methods, especially at lower bit-widths. The table also clearly demonstrates the benefit of using cross-entropy for the last layer instead of the MSE objective.
\begin{table*}[t]
    \centering

    \begin{tabular}{ l | r r r | r r r}
     \toprule
        Model (FP32 accuracy) & \multicolumn{3}{c|}{\rn{18} (69.68)} & \multicolumn{3}{c}{\mnv{2} (71.72)} \\ \midrule
         Bit-width        & \multicolumn{1}{c}{A8}    & \multicolumn{1}{c}{A6}      & \multicolumn{1}{c|}{A4}    & \multicolumn{1}{c}{A8} & \multicolumn{1}{c}{A6}    & \multicolumn{1}{c}{A4} \\\midrule
Min-Max     & 69.60 & 68.19 & 18.82 &	 70.96 & 64.58 & 0.53 \\
MSE         & 69.59 & 67.84 & 31.40 &	 71.35 & 67.55 & 13.57 \\
MSE + Xent  & \textbf{69.60} & \textbf{68.91} & \textbf{59.07} &	\textbf{71.36} & \textbf{68.85 }& \textbf{30.94} \\
BN ($\alpha=6$) & 69.54 & 68.73 & 23.83 &	 71.32 & 65.20 & 0.66 \\
\bottomrule
    \end{tabular}
 \caption{Ablation study for different methods of range setting of (asymmetric uniform) activation quantizers while keeping the weights in FP32. Average ImageNet validation accuracy (\%) over 5 runs.}
\label{tab:rangesetting_activations}
\end{table*}

\subsection{Cross-Layer Equalization}
\label{sec:CLE}
\begin{figure}[t]
    \centering
    \includegraphics[width=0.8\textwidth]{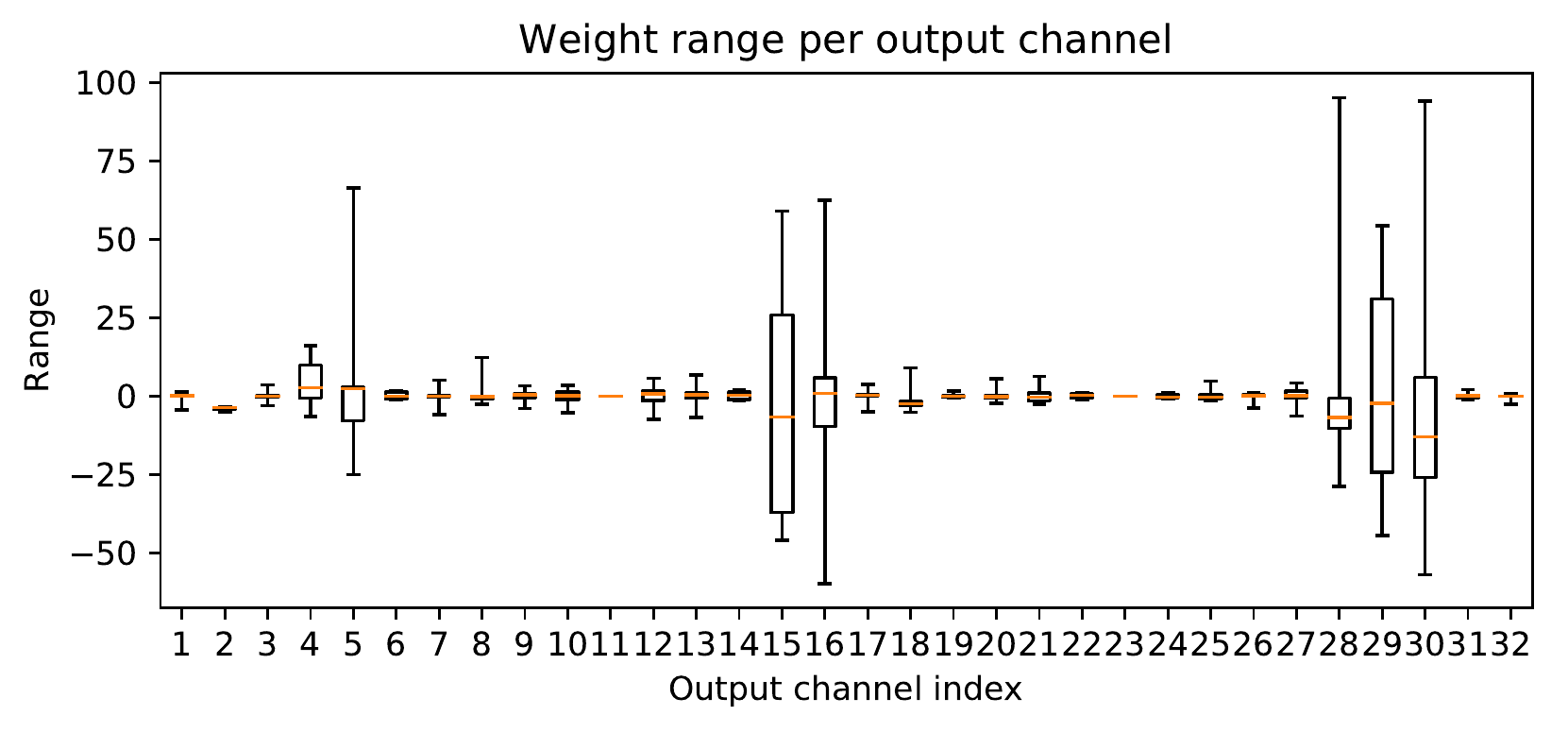}
    \caption{Per (output) channel weight ranges of the first depthwise-separable layer in~\mnv{2} after BN folding. The boxplots show the min and max value, the 2nd and 3rd quartile and the median are plotted for each channel.}
    \label{fig:channel_scales_mnv2}
\end{figure}

A common issue for quantization error is that elements in the same tensor can have significantly different magnitudes. As discussed in the previous section, range setting for the quantization grid tries to find a good trade-off between clipping and rounding error. Unfortunately, in some cases, the difference in magnitude between them is so large that even for moderate quantization (e.g., INT8), we cannot find a suitable trade-off. 
\citet{dfq} showed that this is especially prevalent in depth-wise separable layers since only a few weights are responsible for each output feature and this might result in higher variability of the weights.
Further, they noted that batch normalization folding adds to this effect and can result in a strong imbalance between weights connected to various output channels (see figure \ref{fig:channel_scales_mnv2}). While the latter is less of an issue for a more fine-grained quantization granularity (e.g., per-channel quantization), this remains a big issue for the more widely used per-tensor quantization. Several papers \citep{krishnamoorthi, dfq, sheng2018} noted that efficient models with depth-wise separable convolutions, such as~\mnv{1}~\citep{howard2017mobilenets} and~\mnv{2}~\citep{mobilenetv2}, show a significant drop for PTQ or even result in random performance.

\begin{figure}[t]
    \centering
    \includegraphics[width=0.6\textwidth]{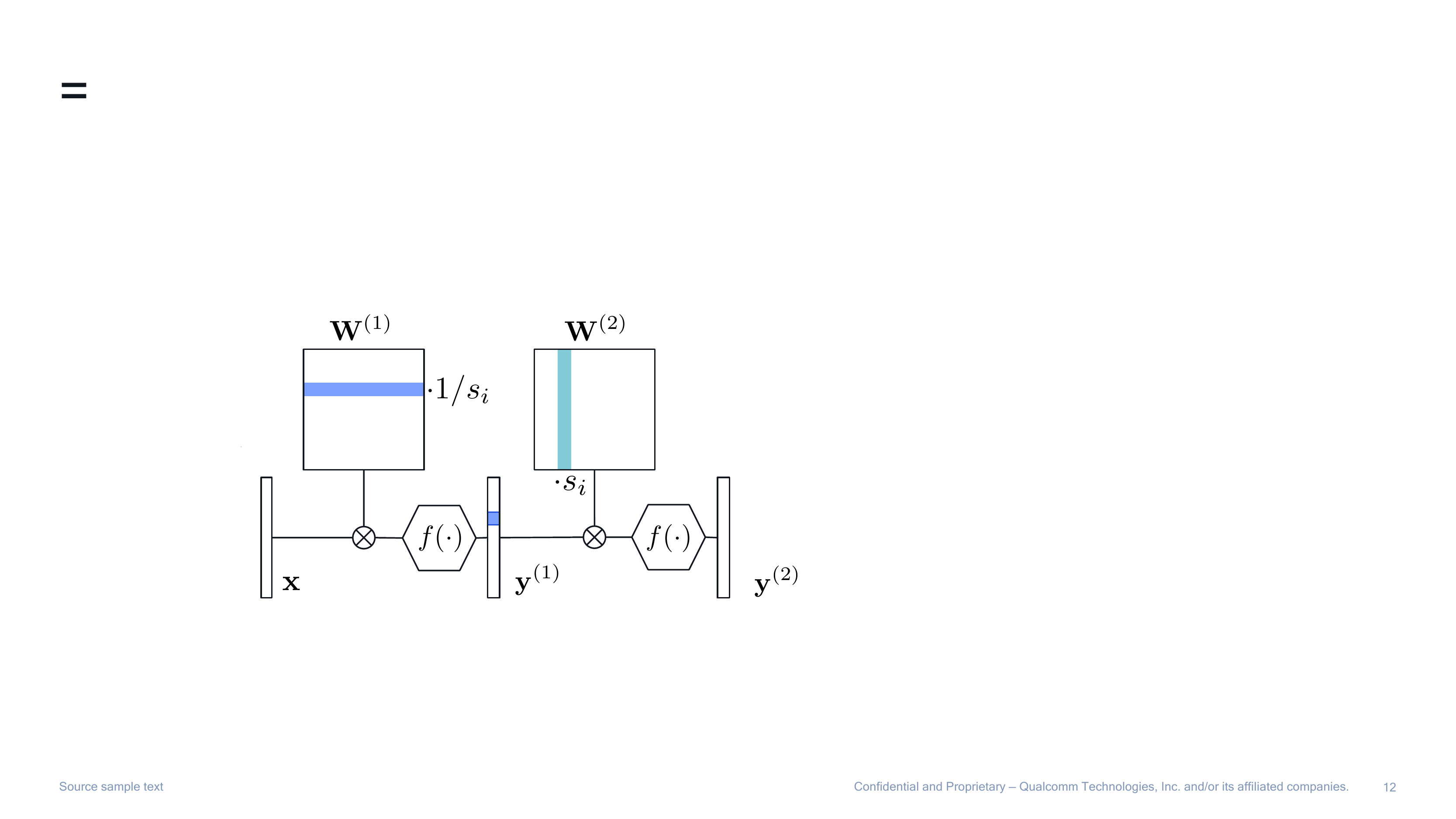}
    \caption{Illustration of the rescaling for a single channel. Scaling a channel in the first layer by a factor $\ct{s}_i$ leads a reparametrization of the equivalent channel in the second layer by $1/\ct{s}_i$.}
    \label{fig:channel_rescaling}
\end{figure}

A solution to overcome such imbalances without the need to use per-channel quantization is introduced by \citet{dfq}. 
A similar approach was introduced in concurrent work by \citet{samesame}. 
In both papers, the authors observe that for many common activation functions (e.g., ReLU, PreLU), a positive scaling equivariance holds:
\begin{equation}
    f(sx) = sf(x)
\label{eq:equivariance}
\end{equation}
for any non-negative real number $s$. This equivariance holds for any homogeneous function of degree one and can be extended to also hold for any piece-wise linear function by scaling its parameterization (e.g. ReLU6).
We can exploit this positive scaling equivariance in consecutive layers in neural networks. Given two layers, $\vec{h} = f(\mati{W}{1} \vec{x} + \veci{b}{1})$ and $\vec{y} = f(\mati{W}{2}\vec{h} + \veci{b}{2})$, through scaling equivariance we have that:
\begin{align}
    \vec{y} &= f(\mati{W}{2} f(\mati{W}{1}\vec{x}+\veci{b}{1}) + \veci{b}{2}) \nonumber \\
        &= f(\mati{W}{2} \mat{S} \hat{f}(\mat{S}^{-1}\mati{W}{1} \vec{x} + \mat{S}^{-1} \veci{b}{1}) + \veci{b}{2}) \nonumber \\
        &= f(\mati{\widetilde{W}}{2} \hat{f}(\mati{\widetilde{W}}{1} \vec{x} + \veci{\widetilde{b}}{1}) + \veci{b}{2})
\end{align}
where $\mat{S}=\diag(\vec{s})$ is a diagonal matrix with value $\mat{S}_{ii}$ denoting the scaling factor $\vec{s}_i$ for neuron $i$. 
This allows us to reparameterize our model with 
$\mati{\widetilde{W}}{2} = \mati{W}{2} \mat{S}$, 
$\mati{\widetilde{W}}{1} = \mat{S}^{-1} \mati{W}{1}$ and 
$\veci{\widetilde{b}}{1} = \mat{S}^{-1} \veci{b}{1}$.
In case of CNNs the scaling will be per-channel and broadcast accordingly over the spatial dimensions. We illustrate this rescaling procedure in figure \ref{fig:channel_rescaling}.

To make the model more robust to quantization, we can find a scaling factor $\vec{s}_i$ such that the quantization noise in the rescaled layers is minimal. 
The \textit{cross-layer equalization} (CLE) procedure \citep{dfq} achieves this by equalizing dynamic ranges across consecutive layers.
They prove that an optimal weight equalization is achieved by setting $\mat{S}$ such that:
\begin{equation}
    \vec{s}_i = \frac{1}{\vec{r}_i^{(2)}}\sqrt{\vec{r}_i^{(1)} \vec{r}_i^{(2)}}
\end{equation}
where $\vec{r}_i^{(j)}$ is the dynamic range of channel $i$ of weight tensor $j$.
The algorithm of~\citet{samesame} introduces a similar scaling factor that also takes the intermediate activation tensor into account. However, they do not have a proof of optimality for this approach.

\paragraph{\textbf{Absorbing high biases}} 
\citet{dfq} further notice that in some cases, especially after CLE, high biases can lead to differences in the dynamic ranges of the activations. Therefore, they propose a procedure to, if possible, absorb high biases into the next layer.
To absorb $\vec{c}$ from layer one (followed by a ReLU activation function $f$) into layer two, we can do the following reparameterization:
\begin{align}
    \vec{y} &= \mati{W}{2} \vec{h} + \veci{b}{2} \nonumber \\
        &= \mati{W}{2} (f(\mati{W}{1} \vec{x} + \veci{b}{1}) + \vec{c}-\vec{c})  + \veci{b}{2} \nonumber \\
        &= \mati{W}{2} (f(\mati{W}{1} \vec{x} + \veci{\widetilde{b}}{1}) + \vec{c})  + \veci{b}{2} \nonumber \\
        &= \mati{W}{2} \vec{\widetilde{h}} + \veci{\widetilde{b}}{2}
        \label{eq:absob_bias}
\end{align}
where $\veci{\widetilde{b}}{2} = \mati{W}{2} \vec{c} + \veci{b}{2}$, $\vec{\widetilde{h}}=\vec{h} - \vec{c}$, and $\veci{\widetilde{b}}{1} = \veci{b}{1}-\vec{c}$. 
In step two, we use the fact that for a layer with ReLU function $f$, there is a non-negative vector $\vec{c}$ such that $r(\mat{W}\vec{x}+\vec{b}-\vec{c}) = r(\mat{W}\vec{x}+\vec{b}) - \vec{c}$.
The trivial solution $\vec{c}=\vec{0}$ holds for all $\vec{x}$.
However, depending on the distribution of $\vec{x}$ and the values of $\vec{W}$ and $\vec{b}$, there can be some values $\vec{c}_i>0$ for which this equality holds for (almost) all $\vec{x}$ in the empirical distribution. 
This value is equal to 
\begin{equation}
    \vec{c}_i = \max\l(0, \min_{\vec{x}} \l(\mat{W}_i^{(1)}\vec{x}+\vec{b}_i^{(1)}\r)\r).
    \label{eq:absob_bias_amount}
\end{equation}
where $\min_{\vec{x}}$ is evaluated on a small calibration dataset. To remove dependence on data, the authors propose to estimate the right hand side of \eqref{eq:absob_bias_amount} by the shift and scale parameters of the batch normalization layer which results\footnote{Assuming $\vec{x}$ is normally distributed, the equality will hold for approximately $99.865\%$ of the inputs.} in $\vec{c}_i=\max(0,\vec{\bnbeta}_i - 3\vec{\bngamma}_i)$. 

\paragraph{\textbf{Experiments}}
In table \ref{tbl:cle_mnv2}, we demonstrate the effect of CLE and bias absorption for quantizing~\mnv{2} to 8-bit. As skip connections break the equivariance between layers, we apply cross-layer equalization only to the layers within each residual block.
Similar to~\citet{krishnamoorthi}, we observe that model performance is close to random when quantizing~\mnv{2} to INT8. Applying CLE brings us back within 2\% of FP32 performance, close to the performance of per-channel quantization. We note that absorbing high biases results in a small drop in FP32 performance, as it is an approximation, but it boosts quantized performance by 1\% due to more precise activation quantization. 
Together, CLE and bias absorption followed by per-tensor quantization yield better results than per-channel quantization.

\begin{table}[t]
     \centering
    \begin{tabular}{ l r r }
      \toprule
         Model          & FP32      & INT8 \\\midrule       
         Original model & 71.72   &  0.12 \\            
         + CLE    & 71.70   & 69.91 \\            
         + absorbing bias & 71.57   & \textbf{70.92}  \\   
         \midrule
         Per-channel quantization & 71.72 & 70.65 \\
         \bottomrule
          \end{tabular}
           \caption{Impact of cross-layer equalization (CLE) for \mnv{2}. ImageNet validation accuracy (\%), evaluated at full precision and 8-bit quantization.}
          \label{tbl:cle_mnv2}
\end{table}

\subsection{Bias correction}
Another common issue is that quantization error is often biased. This means that the expected output of the original and quantized layer or network is shifted ($\eop{\mat{W}\vec{x}}\neq \eop{\matq{W}\vec{x}}$). This kind of  error is more pronounced in depth-wise separable layers with only a few elements per output channel (usually 9 for a $3 \times 3$ kernel). The main contributor to this error is often the clipping error, as a few strongly clipped outliers will likely lead to a shift in the expected distribution. 

Several papers \citep{dfq, samesame, biaswithbias} noted this issue and introduce methods to correct for the expected shift in distribution. 
For a quantized layer $\matq{W}$ with quantization error $\dW = \matq{W} - \mat{W}$, the expected output distribution is
\begin{align}
    \eop{\vecq{y}} &= \eop{\matq{W}\vec{x}} \nonumber  \\
    &= \eop{(\mat{W} + \dW)\vec{x}} \nonumber  \\
    &= \eop{\mat{W}\vec{x}} + \eop{\dW\vec{x}}.
\end{align}
Thus the biased error is given by $\eop{\dW\vec{x}}$. Since $\dW$ is constant, we have that $\eop{\dW\vec{x}} = \dW\eop{\vec{x}}$. 
In case $\dW\eop{\vec{x}}$ is nonzero, the output distribution is shifted.
To counteract this shift we can substract it from the output:
\begin{align}
    \eop{\vec{y}_{\textup{corr}}} &= \eop{\matq{W}\vec{x}} - \dW\eop{\vec{x}} = \eop{\vec{y}}.
    \label{eq:bias_correction}
\end{align}
Note, this correction term is a vector with the same shape as the bias and can thus be absorbed into the bias without any additional overhead at inference time. There are several ways of calculating the bias correction term, the two most common of which are \textit{empirical bias correction} and \textit{analytic bias correction}.

\paragraph{\textbf{Empirical bias correction}}
If we have access to a calibration dataset the bias correction term can simply be calculated by comparing the activations of the quantized and full precision model. In practice, this can be done layer-wise by computing 
\begin{align}
    \dW\eop{\vec{x}} = \eop{\matq{W}\vec{x}} - \eop{\mat{W}\vec{x}}.
\end{align}

\paragraph{\textbf{Analytic bias correction}}
\citet{dfq} introduce a method to analytically calculate the biased error, without the need for data. For common networks with batch normalization and ReLU functions, they use the BN statistics of the preceding layer in order to compute the expected input distribution $\eop{\vec{x}}$. The BN parameters $\vec{\gamma}$ and $\vec{\beta}$ correspond to the mean and standard deviation of the BN layers output. Assuming input values are normally distributied, the effect of ReLU on the distribution can be modeled using the clipped normal distribution. They show that
\begin{align}
    \eop{\vec{x}} &= \eop{\text{ReLU}\left(\vec{x}^{\textup{pre}}\right)} \nonumber \\
    &= \bngamma \mathcal{N}\left( \frac{-\bnbeta}{\bngamma} \right) 
    + \bnbeta \left[\vec{1}-\Phi\left( \frac{-\bnbeta}{\bngamma} \right)\right]
\end{align}
where $\vec{x}^{\textup{pre}}$ is the pre-activation output, which is assumed to be normally distributed with the per-channel means $\bnbeta$ and per-channel standard deviations $\bngamma$, $\Phi(\cdot)$ is the standard normal CDF, and the notation $\mathcal{N}(x)$ is used to denote the standard normal PDF. Note, all vector operations are element-wise (per-channel) operations. After calculating the input distribution $\eop{\vec{x}}$, the correction term can be simply derived by multiplying it with the weight quantization error $\dW$.

\paragraph{\textbf{Experiments}}
In table \ref{tbl:04_bias_correction}, we demonstrate the effect of  bias correction for quantizing~\mnv{2} to 8-bit. Applying analytical bias correction improves quantized model performance from random to over 50\%, indicating that the biased error introduced by quantization significantly harms model performance. When combining bias correction with CLE, we see that both techniques are complementary. Together, they achieve near FP32 performance without using any data.

\begin{table}[t]
    \centering
    \begin{tabular}{l r r}
    \toprule
        Model & FP32 & INT8\\\midrule
        Original Model          & 71.72           & 0.12\\  
        + bias correction               & 71.72           & \textbf{52.02}\\ \midrule
        CLE + bias absorption & 71.57           & 70.92\\   
        + bias correction   & 71.57           & \textbf{71.19}\\ 
     \bottomrule
    \end{tabular}
        \caption{Impact of bias correction for \mnv{2}. ImageNet validation accuracy (\%) evaluated at full precision and 8-bit quantization.}
     \label{tbl:04_bias_correction}
\end{table}

\subsection{AdaRound}
\label{sec:adaround}
\begin{figure}[t]
\centering
  \includegraphics[width=0.6\linewidth]{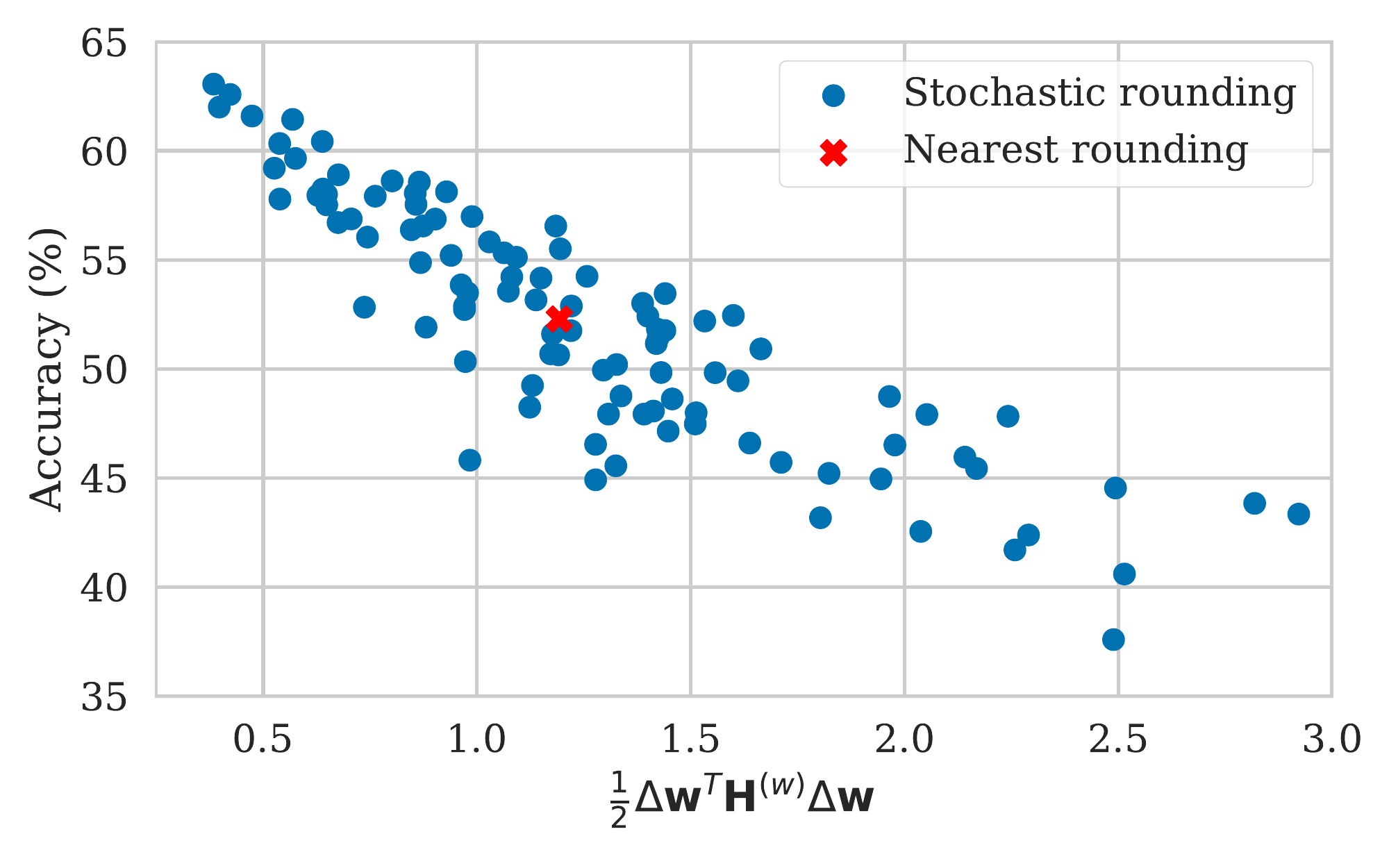}
  \vspace{-0cm}
 \caption{Correlation between the cost in equation~\eqref{eq:adaround_taylor} vs. ImageNet validation accuracy (\%) of 100 stochastic rounding vectors $\vecq{w}$ for $4$-bit quantization of only the first layer of~\rn{18}.}\vspace{-0.1cm}
 \label{fig:adaround_corr}
\end{figure}

Neural network weights are usually quantized by projecting each FP32 value to the \textit{nearest} quantization grid point, as indicated by $\round*{\cdot}$ in equation \eqref{eq:quant_operation} for a uniform quantization grid. We refer to this quantization strategy as rounding-to-nearest. 
The rounding-to-nearest strategy is motivated by the fact that, for a fixed quantization grid, it yields the lowest MSE between the floating-point and quantized weights.
However, \citet{adaround} showed that rounding-to-nearest is not optimal in terms of the task loss when quantizing weights in the post-training regime. To illustrate  this the authors quantized  the weights of the first layer of~\rn{18} to 4 bits using 100 different stochastic rounding samples \citep{Gupta2015} and evaluated the performance of the network for each rounding choice. The best rounding choice among these outperformed rounding-to-nearest by more than $10\%$. Figure~\ref{fig:adaround_corr} illustrates this by plotting the performance of these rounding choices on the y-axis. 
In this section, we describe AdaRound \citep{adaround}, a systematic approach to finding good weight rounding choices for PTQ. AdaRound is a theoretically well-founded and computationally efficient method that shows significant performance improvement in practice.

As the main goal is to minimize the impact of quantization on the final task loss, we start by formulating the optimization problem in terms of this loss
\begin{align}
    \argmin_{\dw}& \quad \eop{\tloss{\vec{x},\vec{y},\vec{w} + \dw} - \tloss{\vec{x},\vec{y},\vec{w}}} \label{eq:adaround_taskloss}
\end{align}
where $\dw$ denotes the perturbation due to quantization and can take two possible values for each weight, one by rounding the weight up  and the other by rounding the weight down. We want to solve this binary optimization problem efficiently. As a first step, we approximate the cost function using a second-order Taylor series expansion. This alleviates the need for performance evaluation for each new rounding choice during the optimization. We further assume that the model has converged, implying that the contribution of the gradient term in the approximation can be ignored, and that the Hessian is block-diagonal, which ignores cross-layer correlations. This leads to the following Hessian based quadratic unconstrained binary optimization (QUBO) problem 
\begin{align}
    \argmin_{\dw^{(\ell)}}& \quad \eop{{\dw^{(\ell)}}^T \mati{H}{\veci{w}{\ell}} \dw^{(\ell)}} \label{eq:adaround_taylor}
\end{align}
The clear correlation in figure~\ref{fig:adaround_corr} between the validation accuracy  and objective of  equation \eqref{eq:adaround_taylor} indicates that the latter serves as a good proxy for the task loss (equation~\ref{eq:adaround_taskloss}),  even for 4-bit weight quantization. 
Despite the performance gains (see table \ref{tab:adaround_ablation}), equation \eqref{eq:adaround_taylor} cannot be widely applied for weight rounding for main two reasons:
\begin{itemize}
    \item The memory and computational complexity of calculating the Hessian is impractical for general use-cases.
    \item The QUBO problem of equation \eqref{eq:adaround_taylor} is NP-Hard. 
\end{itemize}
To tackle the first problem, the authors introduced additional suitable assumptions that allow simplifying the objective of equation~\eqref{eq:adaround_taylor} to the following local optimization problem that minimizes the MSE of the output activations for a layer. 
\begin{align}
   \argmin_{\Delta \mati{W}{\ell}_{k,:}} \quad \eop{\left(\Delta \mati{W}{\ell}_{k,:}
      \veci{x}{\ell-1}\right)^2} \label{eq:adaround_mseopt}
\end{align}
Equation \eqref{eq:adaround_mseopt} requires neither the computation of the Hessian nor any other backward or forward propagation information from the subsequent layers. Note that the approximations and the analysis that have been used to link the QUBO problem of equation \eqref{eq:adaround_taylor} with the local optimization problem of equation \eqref{eq:adaround_mseopt} is independent of the rounding problem. Hence this analysis also benefits the design of algorithms for other problems, including model compression and NAS \citep{BertNAS2020}. 

The optimization of \eqref{eq:adaround_mseopt} is still an NP-hard optimization problem. To find a good approximate solution with reasonable computational complexity, the authors relax the optimization problem to the following continuous optimization problem
\begin{equation}
    \argmin_{\mat{V}} \quad \norm*{\mat{W} \vec{x} - \matsq{W} \vec{x}}^2_F + \lambda \func{f_{\textup{reg}}}{\mat{V}}, \label{eq:adaround_relaxopt}
\end{equation}
where $\norm*{\cdot}^2_F$ denotes the Frobenius norm and $\matsq{W}$ are the soft-quantized weights defined as 
\begin{align}
    \matsq{W} &= \ct{s} \cdot \clamp{ \left(\floor*{\frac{\mat{W}}{\ct{s}}} + \func{h}{\mat{V}}; \ct{n}, \ct{p} \right)}  .\label{eq:adaround_softquant}
\end{align}
We use $\ct{n}$ and $\ct{p}$ to denote integer grid limits, $n=\ct{q}_{\text{min}}/\ct{s}$ and $p=\ct{q}_{\text{max}}/\ct{s}$.
$\mat{V}_{i,j}$ is the continuous variable that we optimize over and $h$ can be any monotonic function with values between $0$ and $1$, i.e.,  $\func{h}{\mat{V}_{i,j}} \in [0, 1]$. In \citet{adaround}, the authors use a rectified sigmoid as $h$. The objective of \eqref{eq:adaround_relaxopt} also introduces a regularizer term that encourages the continuous optimization variables $\func{h}{\mat{V}_{i,j}}$ to converge to either $0$ or $1$, so that they are valid solutions to the discrete optimization in \eqref{eq:adaround_mseopt}. The regularizer used in \citet{adaround} is 
\begin{equation}
    \func{f_{\textup{reg}}}{\mat{V}} = \sum\limits_{i,j} 1 - |2 \func{h}{\mat{V}_{i,j}} - 1|^\beta, \label{eq:adaround_reg}
\end{equation}
where $\beta$ is annealed during the course of optimization to initially allow free movement of $\func{h}{\mat{V}_{i,j}}$ and later to force them to converge to $0$ or $1$. To avoid error accumulation across layers of the neural network and to account for the non-linearity, the authors propose the following final optimization problem
\begin{equation}
    \argmin_{\mat{V}} \norm*{ \func{f_{a}}{\mat{W} \vec{x}} - \func{f_{a}}{\matsq{W} \vec{\hat{x}}}}^2_F + \lambda \func{f_{\textup{reg}}}{\mat{V}}, \label{eq:adaround_asym}
\end{equation}
where $\vec{\hat{x}}$ is the layer's input with all preceding layers quantized and $\funcb{f_a}$ is the activation function. The objective of \eqref{eq:adaround_asym} can be effectively and efficiently optimized using stochastic gradient descent. This approach of optimizing weight rounding is known as AdaRound. 

\begin{table}[t]
    \centering
    \begin{tabular}{ l c c }
        \toprule
         \qquad\qquad\quad Rounding          & First layer      & All layers \\\midrule
         Nearest            &  52.29   &  23.99 \\
         $\mati{H}{\vec{w}}$ task loss (equation~\ref{eq:adaround_taylor}) & 68.62   &  N/A \\
         Cont.~relaxation MSE (equation~\ref{eq:adaround_relaxopt}) &  69.58   &  66.56 \\
         AdaRound (equation~\ref{eq:adaround_asym})     &  \textbf{69.58}   &  \textbf{68.60} \\
         \bottomrule
    \end{tabular}
\caption{Impact of various approximations and assumptions made in section~\ref{sec:adaround} on the ImageNet validation accuracy (\%) for~\rn{18} averaged over 5 runs. N/A implies that the corresponding experiment was computationally infeasible.}
\label{tab:adaround_ablation}
\end{table}

To summarize, the way we round weights during the quantization operation has a significant impact on the performance of the network. AdaRound provides a theoretically sound, computationally fast weight rounding method. It requires only a small amount of unlabeled data samples, no hyperparameter tuning or end-to-end finetuning, and can be applied to fully connected and convolutional layers of any neural network.

\subsection{Standard PTQ pipeline}
\label{sec:standard_ptq}
\begin{figure}
    \centering
    \includegraphics[width=1.0\textwidth]{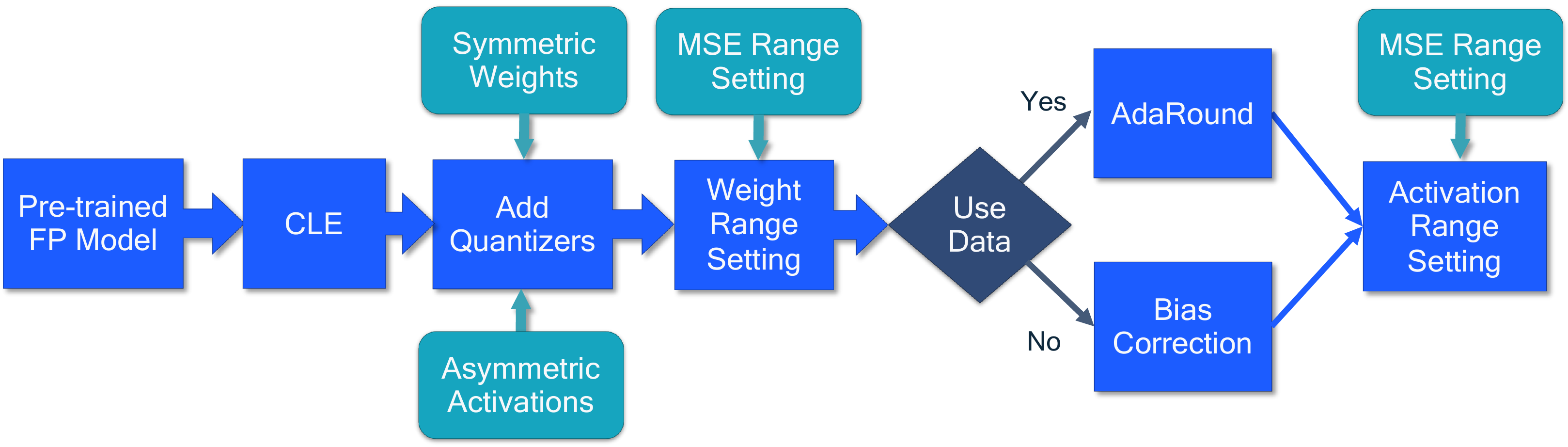}
    \caption{Standard PTQ pipeline. Blue boxes represent required steps and the turquoise boxes recommended choices.}
    \label{fig:standard_ptq}
\end{figure}

In this section, we present a best-practice pipeline for PTQ based on relevant literature and extensive experimentation. We illustrate the recommended pipeline in figure \ref{fig:standard_ptq}. 
This pipeline achieves competitive PTQ results for many computer vision as well as natural language processing models and tasks. Depending on the model, some steps might not be required, or other choices could lead to equal or better performance.
\begin{description}
\item[Cross-layer equalization] First we apply cross-layer equalization (CLE), which is a pre-processing step for the full precision model to make it more quantization friendly. CLE is particularly important for models with depth-wise separable layers and for  per-tensor quantization, but it often also shows improvements for other layers and quantization choices.

\item[Add quantizers] Next we choose our quantizers and add quantization operations in our network as described in section \ref{sec:quantization_simulation}. The choice of quantizer might depend on the specific target HW; for common AI accelerators we recommend using symmetric quantizers for the weights and asymmetric quantizers for the activations. If supported by the HW/SW stack then it is favorable to use per-channel quantization for weights.
\item[Weight range setting] To set the quantization parameters of all weight tensors we recommend using the layer-wise MSE based criteria. In the specific case of per-channel quantization, using the min-max method can be favorable in some cases.
\item[AdaRound] In case we have a small calibration dataset\footnote{Usually, between 500 and 1000 unlabeled images are sufficient as a calibration set.} available we next apply AdaRound in order to optimize the rounding of the weights. This step is crucial to enable low-bit weight quantization (e.g. 4 bits) in the PTQ.
\item[Bias correction] In case we do not have such a calibration dataset and the network uses batch normalization, we can use analytical bias correction instead.
\item[Activation range setting] As the final step, we determine the quantization ranges of all data-dependent tensors in the network (i.e., activations). We use the MSE based criteria for most of the layers, which requires a small calibration set to find the minimum MSE loss. Alternatively, we can use the BN based range setting to have a fully data-free pipeline.
\end{description}

\subsection{Experiments}
\label{sec:ptq_experiments}

\begin{table*}[t]
    \centering
    \begin{tabular}{ l | r| r r |  r r}
    \toprule
                         &       & \multicolumn{2}{c|}{Per-tensor} &
                         \multicolumn{2}{c}{Per-channel} \\ 
        Models        & FP32  & W8A8    & W4A8      & W8A8    & W4A8 \\
        \midrule
\rn{18}	        & 69.68 & 69.60 & 68.62 & 69.56 & 68.91 \\
\rn{50}	        & 76.07 & 75.87 & 75.15 & 75.88 & 75.43 \\
\mnv{2}	        & 71.72 & 70.99 & 69.21 & 71.16 & 69.79 \\
InceptionV3	    & 77.40 & 77.68 & 76.48 & 77.71 & 76.82 \\
EfficientNet lite   & 75.42 & 75.25 & 71.24 & 75.39 & 74.01 \\
DeeplabV3	        & 72.94 & 72.44 & 70.80 & 72.27 & 71.67 \\
EfficientDet-D1	    & 40.08 & 38.29 & 0.31 & 38.67 & 35.08\\
BERT-base\textsuperscript{\dag} & 83.06 & 82.43 & 81.76 & 82.77 & 82.02 \\
\bottomrule
\end{tabular}
 \caption{Performance (average over 5 runs) of our standard PTQ pipeline for various models and tasks. DeeplabV3 (MobileNetV2 backbone) is evaluated on Pascal VOC (mean intersection over union), EfficientDet-D1 on COCO 2017 (mean average precision), BERT-base on the GLUE benchmark and other models on ImageNet (accuracy). We evaluate all models on the respective validation sets. Higher is better in all cases.~\textsuperscript{\dag}A few quantized activations are kept in higher precision (16 bits).}
 \label{tbl:standard_ptq}
\end{table*}


We now evaluate the performance of the aforementioned PTQ pipeline on common computer vision and natural language understanding applications. Our results are summarized in table \ref{tbl:standard_ptq}.
For the task of semantic segmentation, we evaluate DeepLabV3 (with a \mnv{2} backbone) \citep{deeplabv3} on Pascal VOC and for object detection, EfficientDet \citep{tan2020efficientdet} on COCO 2017. The rest of the computer vision models are evaluated on the ImageNet classification benchmark. For natural language understanding, we evaluate BERT-base on the GLUE benchmark \citep{wang2019glue}.

In all cases, we observe that 8-bit quantization of weights and activation (W8A8) leads to only marginal loss of accuracy compared to floating-point (within 0.7\%) for all models. For W8A8 quantization we also see no significant gains from using per-channel quantization. However, the picture changes when weights are quantized to 4 bits (W4A8).
For ResNet18/50 and InceptionV3 the accuracy drop is still within 1\% of floating-point for both per-tensor and per-channel quantization. 
However, for more efficient networks, such as \mnv{2} and EfficientNet lite, the drop increases to 2.5\% and 4.2\% respectively for per-tensor quantization. This is likely due to the quantization of the depth-wise separable convolutions. 
Here, per-channel quantization can show a significant benefit, for example, in EfficientNet lite per-channel quantization increases the accuracy by 2.8\% compared to per-tensor quantization, bringing it within 1.4\% of full-precision accuracy. We see similar effects for EfficientDet-D1 and DeeplabV3 which both uses depth-wise separable convolutions in their backbone. 

For BERT-base, we observe that a few activation tensors have extreme differences in their dynamic ranges. To make PTQ still work, we identified these layers using our debugging procedure outlined in section \ref{sec:debugging} and kept them in 16 bit.
Otherwise BERT-base follows similar trends as most other models and our PTQ pipeline allows 4 bit weight quantization within 1.5\% drop in GLUE score.

\subsection{Debugging}
\label{sec:debugging}

\begin{figure}[p]
    \centering
    \includegraphics[width=\textwidth]{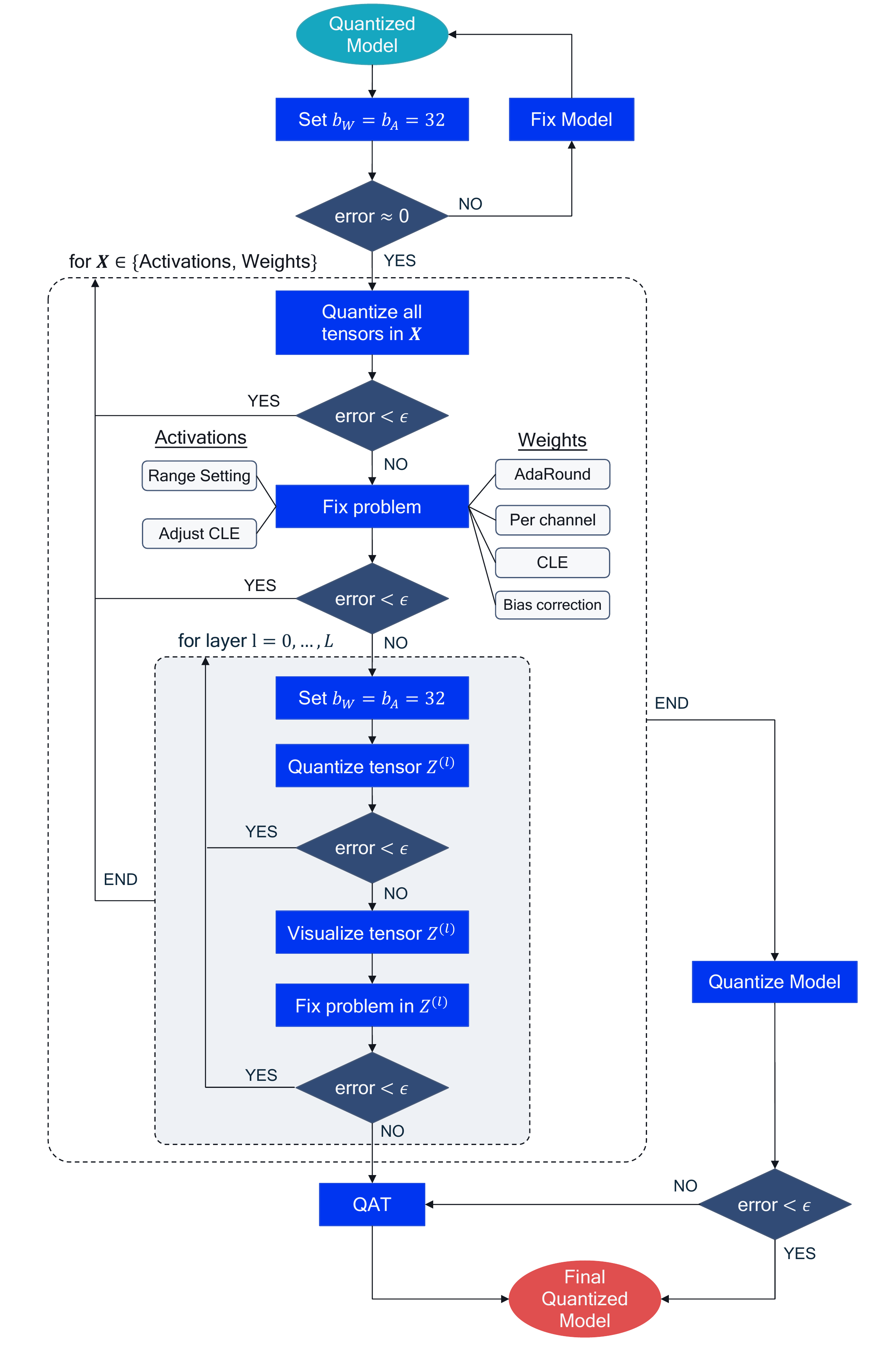}
    \caption{PTQ debugging flow chart. Error is the difference between floating-point and quantized model accuracy. }
    \label{fig:ptw_debugging_flowchart}
\end{figure}

We showed that the standard PTQ pipeline can achieve competitive results for a wide range of models and networks. However, if after following the steps of our pipeline, the model's performance is still not satisfactory, we recommend a set of diagnostics steps to identify the bottlenecks and improve the performance. While this is not strictly an algorithm, these debugging steps can provide insights on why a quantized model underperforms and help to tackle the underlying issues. These steps are shown as a flow chart in figure \ref{fig:ptw_debugging_flowchart} and are described in more detail below:

\begin{description}
    \item[FP32 sanity check] An important initial debugging step is to ensure that the floating-point and quantized model behave similarly in the forward pass, especially when using custom quantization pipelines. Set the quantized model bit-width to 32 bits for both weights and activation, or by-pass the quantization operation, if possible, and check that the accuracy matches that of the FP32 model.
    
    \item[Weights or activations quantization] The next debugging step is to identify how activation or  weight quantization impact the performance independently. Does performance recover if all weights are quantized to a higher bit-width while activations are kept in a lower bit-width, or conversely if all activations use a high bit-width and activations a low bit-width? This step can show the relative contribution of activations and weight quantization to the overall performance drop and point us towards the appropriate solution.
    
    \item[Fixing weight quantization] If the previous step shows that weight quantization does cause significant accuracy drop, then there are a few solutions to try:
    \begin{itemize}
        \item Apply CLE if not already implemented, especially for models with depth-wise separable convolutions. 
        \item Try per-channel quantization. This will address the issue of uneven per-channel weight distribution.
        \item Apply bias correction or AdaRound if calibration data is available.
    \end{itemize}
     \item[Fixing activation quantization] To reduce the quantization error from activation quantization, we can also try using different range setting methods or adjust CLE to take activation quantization ranges into account, as vanilla CLE can lead to uneven activation distribution.
    \item[Per-layer analysis] If the global solutions have not restored accuracy to acceptable levels, we consider each quantizer individually. We set each quantizer sequentially, to the target bit-width while keeping the rest of the network to 32 bits (see inner for loop in figure \ref{fig:ptw_debugging_flowchart}). 
    
    \item[Visualizing layers] If the quantization of a individual tensor leads to significant accuracy drop, we recommended visualizing the tensor distribution at different granularities, e.g. per-channel as in figure \ref{fig:channel_scales_mnv2}, and dimensions, e.g., per-token or per-embedding for activations in BERT.
 
    \item[Fixing individual quantizers] The visualization step can reveal the source of the tensor's sensitivity to quantization. Some common solutions involve custom range setting for this quantizer or allowing a higher bit-width for problematic quantizer, e.g., BERT-base from table~\ref{tbl:standard_ptq}. If the problem is fixed and the accuracy recovers, we continue to the next quantizer. If not, we may have to resort to other methods, such as quantization-aware training (QAT), which is discussed in section \ref{sec:QAT}.
\end{description}

After completing the above steps, the last step is to quantize the complete model to the desired bit-width. If the accuracy is acceptable, we have our final quantized model ready to use. Otherwise, we can consider higher bit-widths and smaller granularities or revert to more powerful quantization methods, such as quantization-aware training.

\section{Quantization-aware training}
\label{sec:QAT}
The post-training quantization techniques described in the previous section  are the first go-to tool in our quantization toolkit. They are very effective and fast to implement because they do not require retraining of the network with labeled data. However, they have limitations, especially when aiming for low-bit quantization of activations, such as 4-bit and below. Post-training techniques may not be enough to mitigate the large quantization error incurred by low-bit quantization. In these cases, we resort to \textit{quantization-aware training} (QAT). QAT models the quantization noise source (see section \ref{sec:quantization_simulation}) during training. This allows the model to find more optimal solutions than post-training quantization. However, the higher accuracy comes with the usual costs of neural network training, i.e., longer training times, need for labeled data and hyper-parameter search.

In this section, we explore how back-propagation works in networks with simulated quantization and provide a standard pipeline for training models with QAT effectively.
We will also discuss the implications of batch normalization folding and per-channel quantization in QAT and provide results for a wide range of tasks and models.

\subsection{Simulating quantization for backward path}
\label{sec:Simulating Quantization For Backwards Path}
\begin{figure}[h]
    \centering
    \includegraphics[width=0.5\textwidth]{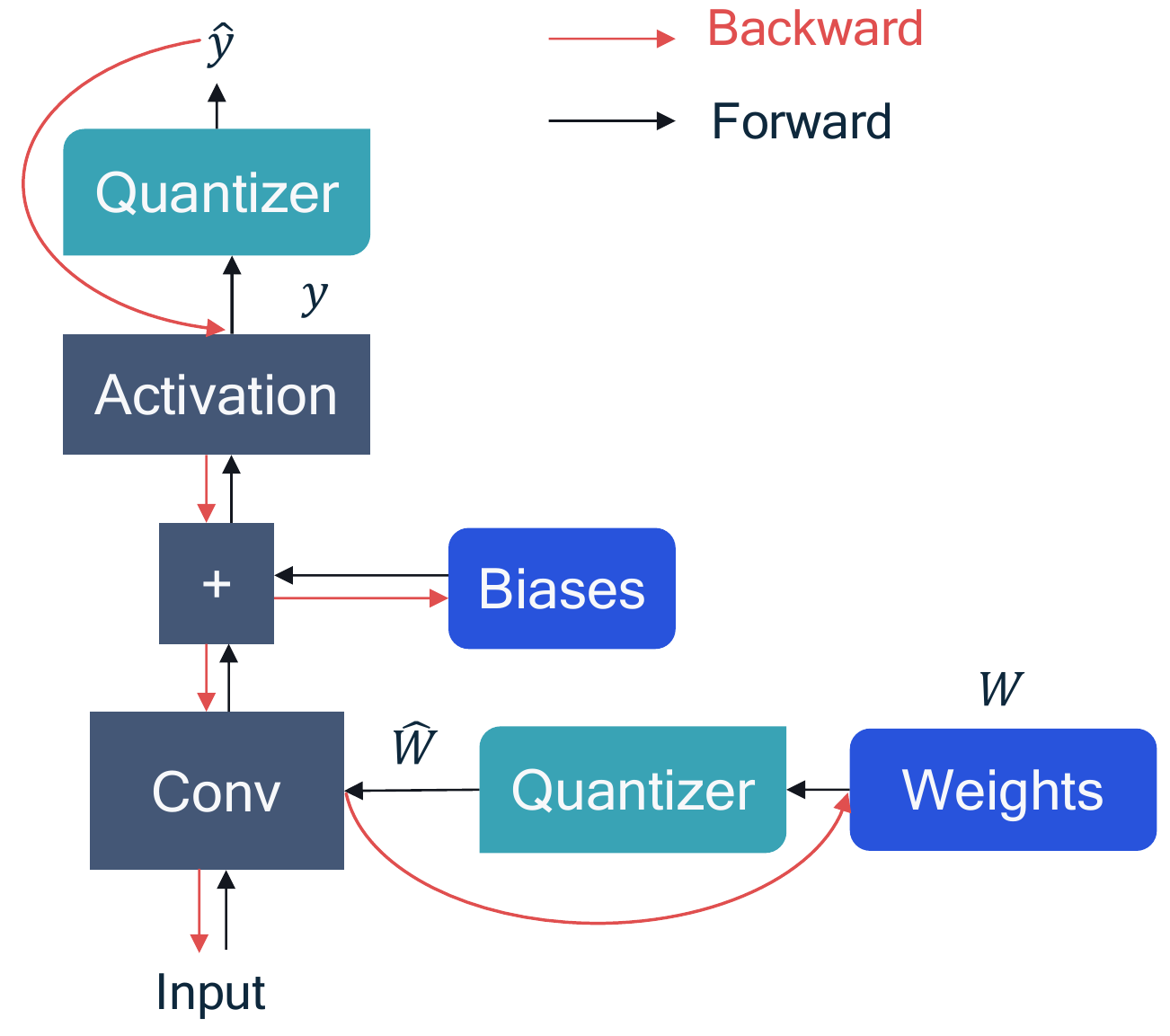}
    \caption{Forward and backward computation graph for quantization aware training with STE assumption.}
    \label{fig:sim_quant_backward}
\end{figure}

In section \ref{sec:quantization_simulation}, we saw how quantization can be simulated using floating-point in deep learning frameworks. However, if we look at the computational graph of figure \ref{fig:simulation}, to train such a network we need to back-propagate through the simulated quantizer block. This poses an issue because the gradient of the round-to-nearest operation in equation \eqref{eq:quant_operation} is either zero or undefined everywhere, which makes gradient-based training impossible. A way around this would be to approximate the gradient using the \textit{straight-through estimator} (STE,~\citealt{bengio2013estimating}), which approximates the gradient of the rounding operator as $1$:
\begin{equation}
\label{eq:STE_round}
    \frac{\partial \round*{\ct{y}}}{\partial \ct{y}} = 1 
\end{equation}
Using this approximation we can now calculate the gradient of the quantization operation from equation \eqref{eq:quant_function}. For clarity we assume symmetric quantization, namely $\ct{z}=0$, but the same result applies to asymmetric quantization since the zero-point is a constant. We use $\ct{n}$ and $\ct{p}$ to define the integer grid limits, such that $n=\ct{q}_{\text{min}}/\ct{s}$ and $p=\ct{q}_{\text{max}}/\ct{s}$. The gradient of equation \eqref{eq:quant_function} \textit{w.r.t} its input, $\vec{x}_i$, is given by:
\begin{align}
    \frac{\partial \widehat{\vec{x}}_i}{\partial \vec{x}_i} &= \frac{\partial q( \vec{x}_i)}{\partial \vec{x}_i} \nonumber\\
    & =s \cdot \frac{\partial}{\partial \vec{x}_i} \clamp{\l( \round*{\frac{\vec{x}_i}{\ct{s}}} ; \ct{n},\ct{p}\r)} + 0\nonumber\\
    & =\begin{dcases}
       \ct{s} \cdot \frac{\partial \round*{\vec{x}_i/\ct{s}}}{\partial(\vec{x}_i/\ct{s})}  \frac{\partial (\vec{x}_i/\ct{s})}{\partial \vec{x}_i} &  \text{if } \ct{q}_{\text{min}} \leq \vec{x}_i\leq \ct{q}_{\text{max}},\\
        \ct{s} \cdot \frac{\partial \ct{n} }{\partial \vec{x}_i} & \text{if } \vec{x}_i < \ct{q}_{\text{min}}, \\
         \ct{s} \cdot \frac{\partial \ct{p} }{\partial \vec{x}_i} & \text{if } \vec{x}_i > \ct{q}_{\text{max}},
       \end{dcases} 
       \nonumber\\
        & =\begin{dcases}
        1  &  \text{if }  \ct{q}_{\text{min}} \leq \vec{x}_i \leq \ct{q}_{\text{max}}, \\
        0 &  \text{otherwise}.
        \end{dcases}\label{eq:ste_quant_operation}
\end{align}
Using this gradient definition we can now back-propagate through the quantization blocks. Figure \ref{fig:sim_quant_backward} shows a simple computational graph for the forward and backward pass used in quantization-aware training. The forward pass is identical to that of figure~\ref{fig:simulation}, but in the backward pass we effectively skip the quantizer block due to the STE assumption. 
In earlier QAT work the quantization ranges for weights and activations were updated at each iteration most commonly using the min-max range \citep{krishnamoorthi}. In later work \citep{lsq,tqt,lsq+}, the STE is used to calculate the gradient \textit{w.r.t.} the quantization parameters, $\ct{z}$ and $\ct{s}$. Using the chain rule and the STE, we first calculate the gradient \textit{w.r.t.} the scale-factor:
\begin{align}
       \frac{\partial \widehat{\vec{x}}_i}{\partial \ct{s}}  &=\frac{\partial} {\partial \ct{s}} \l[\ct{s} \cdot \clamp{\l(\round*{\frac{\vec{x}_i}{s}} ; n, p \r)}\r]  \nonumber\\
       & = \begin{cases}
                -\vec{x}_i/\ct{s} + \round*{\vec{x}_i/\ct{s}}  &  \text{if }  \ct{q}_{\text{min}} \leq \vec{x}_i\leq \ct{q}_{\text{max}}, \\
               \ct{n} &  \text{if } \vec{x}_i< \ct{q}_{\text{min}},\\
               \ct{p} &   \text{if } \vec{x}_i> \ct{q}_{\text{max}}.  \\
        \end{cases} \label{eq:grad_scale_factor}
\end{align}
Originally, we restricted the zero-point to be an integer. To make zero-point learnable we convert into a real number and apply the rounding operator. The modified quantization function is defined as:
\begin{equation}
    \label{eq:quant_function_learned_asymmetric}
    \widehat{\vec{x}} = q(\vec{x}; \ct{s},\ct{z}) = \ct{s}\cdot \l[ \clamp{\l(\round*{\frac{\vec{x}}{s}} + \round*{z}; n, p \r)} -  \round*{z} \r]
\end{equation}
The gradient \textit{w.r.t.} to $\ct{z}$ is calculated by applying the STE once again to the rounding operator:
\begin{equation}
    \label{eq:zero_point_grad}
    \frac{\partial \widehat{\vec{x}}_i}{\partial z}= 
    \begin{cases}
          0 & \ct{q}_{\text{min}} \leq \vec{x}_i\leq \ct{q}_{\text{max}},  \\
          -s &  \text{otherwise}.
    \end{cases}
\end{equation}

\subsection{Batch normalization folding and QAT} \label{sec:qat_bn_folding}
In section \ref{sec:batch_norm_folding}, we introduced batch normalization folding that absorbs the scaling and addition into a linear layer to allow for more efficient inference. During quantization-aware training, we want to simulate inference behavior closely, which is why we have to account for BN-folding during training. Note that in some QAT literature, the BN-folding effect is ignored. While this is fine when we employ \textit{per-channel} quantization (more below in this section), keeping BN unfolded for per-tensor quantization will result in one of the two following cases:
\begin{enumerate}
     \item The BN layer applies per-channel rescaling during inference. In this case we might as well use per-channel quantization in the first place.
    \item We fold BN during deployment into the weight tensor and incur potentially significant accuracy drop as we trained the network to adapt to a different quantization noise.
\end{enumerate}


A simple but effective approach to modeling BN-folding in QAT is to \textit{statically fold} the BN scale and offset into the linear layer's weights and bias, as we saw in equations~\eqref{eq:folded_weights} and~\eqref{eq:folded_bias}. This corresponds to re-parametrization of the weights and effectively removes the batch normalization operation from the network entirely. When starting from a converged pre-trained model, static folding is very effective, as we can see from the result of table \ref{tbl:bn_ablation}. 

An alternative approach by \citet{jacob2018cvpr} both updates the running statistics during QAT and applies BN-folding using a correction. This approach is more cumbersome and computationally costly because it involves a \textit{double forward pass}: one for the batch-statistics and one for the quantized linear operation. However, based on our experiments (see table \ref{tbl:bn_ablation}), static-folding performs on par or better despite its simplicity. 

\begin{table*}[t]
    \centering
    \begin{tabular}{ l | r r | r r}
    \toprule
        Model (FP32 accuracy)                   & \multicolumn{2}{c|}{\rn{18} (69.68)} & \multicolumn{2}{c}{\mnv{2} (71.72)} \\ 
        \midrule
        Bit-width                           & W4A8  & W4A4   & W4A8   & W4A4 \\\midrule
        Static folding BN                   & \textbf{69.76} & \textbf{68.32}  & \textbf{70.17}  & \textbf{66.43} \\
        Double forward~\citep{krishnamoorthi} & 69.42 & 68.20  & 66.87  & 63.54 \\ \midrule
        Static folding (per-channel)        & 69.58 & 68.15  & \textbf{70.52}  & 66.32 \\ 
        Keep original BN (per-channel)      & \textbf{70.01} & \textbf{68.83}  & 70.48  & \textbf{66.89} \\
        \bottomrule
    \end{tabular}
     \caption{Ablation study with various ways to include BN into QAT. The learning rate is individually optimized for each configuration. Average ImageNet validation accuracy (\%) over 3 runs.}
     \label{tbl:bn_ablation}
\end{table*}

\paragraph{\textbf{Per-channel quantization}}
In section \ref{sec:per_channel_quantization}, we mentioned that per-channel quantization of the weights can improve accuracy when it is supported by hardware. The static folding re-parametrization is also valid for per-channel quantization. However, per-channel quantization provides additional flexibility as it allows us to absorb the batch normalization scaling operation into the per-channel scale-factor. Let us see how this is possible by revisiting the BN folding equation from section~\ref{sec:batch_norm_folding}, but this time introduce per-channel quantization of the weights, such that $\widehat{\mat{W}}_{k,:}=q(\mat{W}_{k, :} \,; \ct{s}
_{\vec{w},k})= \ct{s}_{\vec{w},k} \mat{W}^{\text{int}}_{k,:}$. By applying batch normalization to the output of a linear layer similar to equation \eqref{eq:bn_folding}, we get:
\begin{align}
\label{eq:per_channel_BN}
    \widehat{\vec{y}}_k & = \BatchNorm\l(\widehat{\mat{W}}_{k,:} \, \vec{x}\r) \nonumber \\
  & =\frac{\bngamma_k \widehat{\mat{W}}_{k,:}}{\sqrt{\bnsigma_k^2 + \epsilon}}  \vec{x} + \l(\bnbeta_k - \frac{\bngamma_k \bnmu_k}{\sqrt{\bnsigma_k^2 + \epsilon}}\r) \nonumber\\
  &= \frac{\bngamma_k s_{\vec{w},k}}{\sqrt{\bnsigma_k^2 + \epsilon}}  \mat{W}^{\text{int}}_{k,:} \, \vec{x} + \widetilde{\vec{b}}_k \nonumber \\
  & = \widetilde{\ct{s}}_{\vec{w},k} \l(\mat{W}^{\text{int}}_{k,:} \, \vec{x} \r) + \widetilde{\vec{b}}_k
\end{align}

We can see that it is now possible to absorb the batch normalization scaling parameters into the per-channel scale-factor. 
For QAT, this means that we can keep the BN layer intact during training and merge the BN scaling factor into the per-channel quantization parameters afterward.
In practice, this modeling approach is on par or better for per-channel quantization compared to static folding as we can see from the last two rows of table \ref{tbl:bn_ablation}.

\subsection{Initialization for QAT}
In this section, we will explore the effect of initialization for QAT. It is common practice in literature to start from a pre-trained FP32 model \citep{lsq,krishnamoorthi,jacob2018cvpr,tqt}. While it is clear that starting from an FP32 model is beneficial, the effect of the quantization initialization on the final QAT result is less studied. Here we explore the effect of using several of our PTQ techniques as an initial step before doing QAT.

\paragraph{\textbf{Effect of range estimation}}
To assess the effect of the initial range setting (see section \ref{sec:range_setting}) for weights and activations, we perform two sets of experiments, which are summarized in table \ref{tbl:quant_init_ablation}. In the first experiment, we quantize the weights to 4-bits  and keep the activations in 8-bits. We compare the min-max initialization with the MSE based initialization for the weights quantization range. While the MSE initialized model has a significantly higher starting accuracy, the gap closes after training for 20 epochs.

\begin{figure}[h]
    \centering
    \includegraphics[width=0.7\textwidth]{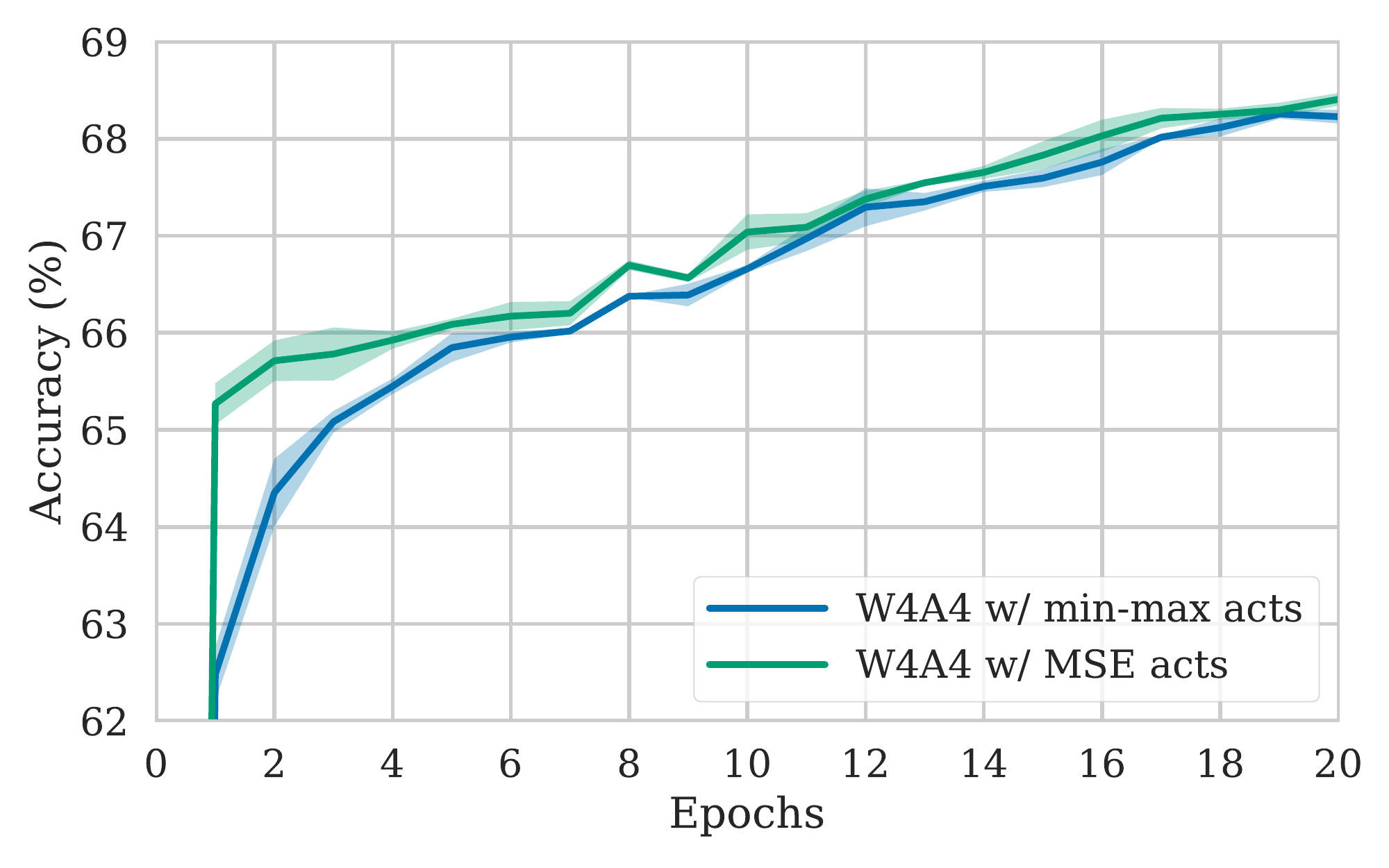}
    \vspace{-0.2cm}
    \caption{Influence of the initial activation range setting on the QAT training behavior of \rn{18}. Average ImageNet validation accuracy (\%) after each training epoch over 3  runs (and standard deviation shaded).}
    \label{fig:act_quant_init}
\end{figure}

To explore the same effect for activation quantization, we perform a similar experiment, where we now quantize the activation to 4-bits and compare min-max initialization with MSE based initialization.
The observations from weight range initialization hold here as well. In figure \ref{fig:act_quant_init} we show the full training curve of this experiment. In the first few epochs, there is a significant advantage for using MSE initialization, which almost vanishes in the later stage of training. In conclusion, a better initialization can lead to better QAT results, but the gain is usually small and vanishes the longer the training lasts.

\begin{table*}[t]
    \centering
    \begin{tabular}{ l | r r | r r}
    \toprule
        Model (FP32 accuracy)                   & \multicolumn{2}{c|}{\rn{18} (69.68)} & \multicolumn{2}{c}{\mnv{2} (71.72)} \\ 
                                            & PTQ   & QAT    &  PTQ   & QAT \\ \midrule
        W4A8 w/ min-max weight init         &  0.12 &  69.61 & 0.56 & 69.96  \\
        W4A8 w/ MSE weight init             &  18.58 &  \textbf{69.76} & 12.99 & \textbf{70.13}  \\ \midrule
        W4A4 w/ min-max act init            &  7.51 & 68.23 & 0.22 & \textbf{66.55} \\ 
        W4A4 w/ MSE  act init               &  9.62 & \textbf{68.41} & 0.71 & 66.29 \\ 
        \bottomrule
    \end{tabular}
     \caption{Ablation study for various ways to initialize the quantization grid. The learning rate is individually optimized for each configuration. ImageNet validation accuracy (\%) averaged over 3 runs.}
 \label{tbl:quant_init_ablation}
\end{table*}

\paragraph{\textbf{Effect of CLE}}
In table \ref{tbl:cle_init_ablation} we compare the effect of other PTQ improvements such as CLE and bias correction. While for~\rn{18} we do not see a significant difference in the final QAT performance, for~\mnv{2} we observe that it cannot be trained without CLE. This is likely due to the catastrophic performance drop caused by per-tensor quantization, which we discussed in section \ref{sec:CLE}.

In conclusion, for models that have severe issues with plain PTQ we may need advanced PTQ techniques such as CLE to initialize QAT. In most other cases, an improved PTQ initialization leads only to a minor improvement in the final QAT performance.

\begin{table*}[t]
    \centering
    \begin{tabular}{ l | r r | r r}
    \toprule
        Model (FP32 accuracy)                   & \multicolumn{2}{c|}{\rn{18} (69.68)} & \multicolumn{2}{c}{\mnv{2} (71.72)} \\ 
                                            & PTQ   & QAT    &  PTQ   & QAT \\ \midrule
        W4A8 baseline                       &  18.58 & 69.74  & 0.10 & 0.10 \\ 
        W4A8 w/ CLE                         &  16.29 & \textbf{69.76}  & 12.99 & \textbf{70.13} \\ 
        W4A8 w/ CLE + BC                    & 38.58  &  69.72 & 46.90 & 70.07 \\
        \bottomrule
    \end{tabular}
    \caption{Ablation study with various PTQ initialization. The learning rate is individually optimized for each configuration. ImageNet validation accuracy (\%) averaged over 3 runs.}
     \label{tbl:cle_init_ablation}
\end{table*}


\subsection{Standard QAT pipeline}
\begin{figure}[h]
    \centering
    \includegraphics[width=1.0\textwidth]{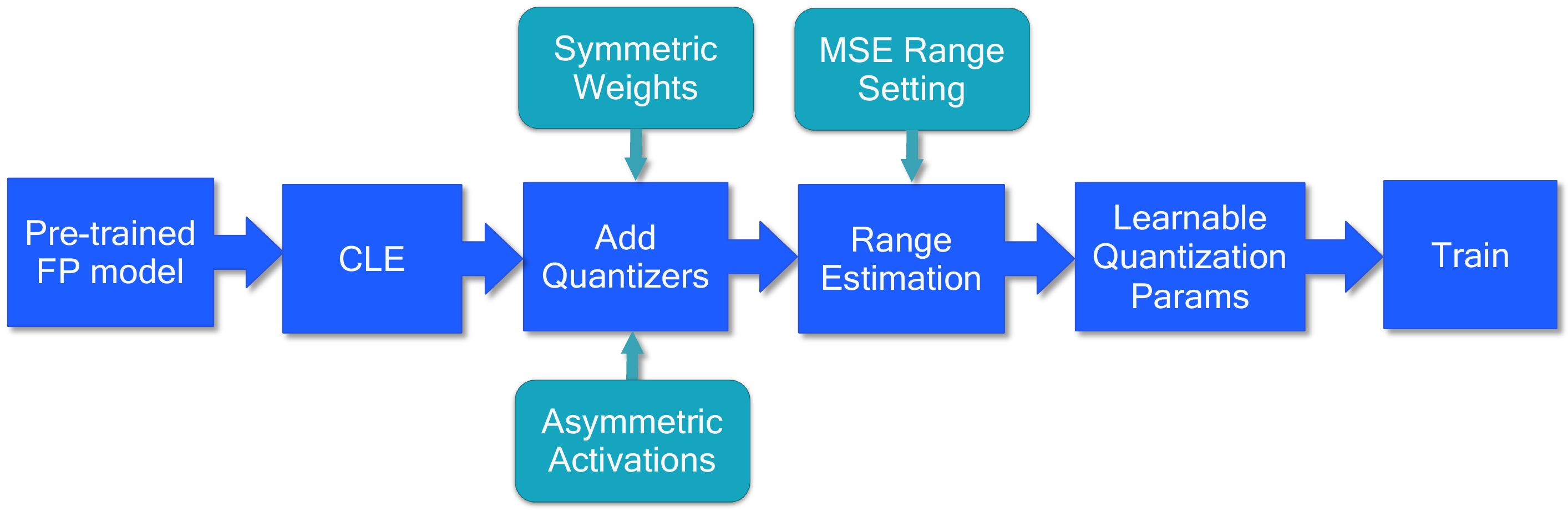}
    \caption{Standard quantization-aware training pipeline. The blue boxes represent the steps and the turquoise boxes recommended choices.}
    \label{fig:qat_pipeline}
\end{figure}

In this section, we present a best-practice pipeline for QAT based on relevant literature and extensive experimentation. We illustrate the recommended pipeline in figure \ref{fig:qat_pipeline}. This pipeline yields good  QAT results over a variety of computer vision and natural language processing models and tasks, and can be seen as the go-to tool for achieving low-bit quantization performance. As discussed in previous sections, we always start from a pre-trained model and follow some PTQ steps in order to have faster convergence and higher accuracy.

\begin{description}
\item[Cross-layer equalization] Similar to PTQ, we first apply CLE to the full precision model. As we saw in table \ref{tbl:cle_init_ablation}, this step is necessary for models that suffer from imbalanced weight distributions, such as MobileNet architectures. For other networks or in the case of per-channel quantization this step can be optional.

\item[Add quantizers] Next, we choose our quantizers and add quantization operations in our network as described in section \ref{sec:quantization_simulation}. The choice for quantizer might depend on the specific target HW, for common AI accelerators we recommend using symmetric quantizers for the weights and asymmetric quantizers for the activations. If supported by the HW/SW stack, then it is favorable to use per-channel quantization for weights. At this stage we will also take care that our simulation of batch normalization is correct, as discussed in section \ref{sec:qat_bn_folding}.

\item[Range estimation] Before training we have to initialize all quantization parameters. A better initialization will help faster training and might improve the final accuracy, though often the improvement is small (see table \ref{tbl:quant_init_ablation}).
In general, we recommend to set all quantization parameters using the layer-wise MSE based criteria. In the specific case of per-channel quantization, using the min-max setting can sometimes be favorable.

\item[Learnable Quantization Parameters]
We recommend making the quantizer paramaters learnable, as discussed in section \ref{sec:Simulating Quantization For Backwards Path}. Learning the quantization parameters directly, rather than updating them at every epoch, leads to higher performance especially when dealing with low-bit quantization.  However, using learnable quantizers requires special care when setting up the optimizer for the task. When using SGD-type optimizers, the learning rate for the quantization parameters needs to be reduced compared to the rest of the network parameters. The learning rate adjustment can be avoided if we use optimizers with adaptive learning rates such as Adam or RMSProp.
\end{description}

\subsection{Experiments}
\label{sec:QAT_Results}
Using our QAT pipeline, we quantize and evaluate the same models we used for PTQ in section \ref{sec:ptq_experiments}. Our results are presented in table \ref{tbl:standard_qat} for different bit-widths and quantization granularities. DeepLabV3 is trained for 80 epochs on Pascal VOC; EfficientDet for 20 epochs on COCO 2017; all other vision models are trained for 20 epochs on ImageNet. BERT-base is trained on each of the corresponding GLUE tasks for 3 to 12 epochs depending on the task and the quantization granularity. We use the Adam optimizer for all models. We present the results with the best learning rate per quantization configuration and perform no further hyper-parameter tuning.

We observe that for networks without depth-wise separable convolutions (first 3 rows of table \ref{tbl:standard_qat}), W8A8 and W4A8 quantization perform on par with and even outperform the floating-point model in certain cases. This could be due to the regularizing effect of training with quantization noise or due to the additional fine-tuning during QAT. For the more aggressive W4A4 case, we notice a small drop but still within 1\% of the floating-point accuracy.

Quantizing networks with depth-wise separable layers (\mnv{2}, EfficientNet lite, DeeplabV3, EfficientDet-D1) is more challenging; a trend we also observed from the PTQ results in section \ref{sec:ptq_experiments} and discussed in the literature \citep{OneBitwidth, depthsep_friendly_quant}. Whereas 8-bit quantization incurs close to no accuracy drop, quantizing weights to 4 bits leads to a larger drop, e.g. approximately 4\% drop for EfficientNet lite with per-tensor quantization. Per-channel quantization can improve performance significantly bringing DeepLabV3 to floating-point accuracy and reducing the gap of MobileNetV2 and EfficientNet lite to less than 1.5\%. 
Quantizing both weights and activations to 4-bits remains a challenging for such networks, even with per-channel quantization it can lead to a drop of up to 5\%. EfficientDet-D1 remains more difficult to quantize than the other networks in this group.

For BERT-base we observe that QAT with range learning can efficiently deal with the high dynamic ranges allowing to keep all activations in 8 bits (unlike for PTQ). W4A8 stays within 1\% of the original GLUE score, indicating that low bit weight quantization is not a problem for transformer models. We only notice a significant drop in performance when combining this with low bit activation quantization (W4A4).
 \begin{table*}[t]
    \centering
    \begin{tabular}{ l | r| r r r |  r r r}
    \toprule
                         &       & \multicolumn{3}{c|}{Per-tensor} & \multicolumn{3}{c}{Per-channel} \\ 
        Models        & FP32  & W8A8    & W4A8      & W4A4  &  W8A8    & W4A8      & W4A4 \\
        \midrule
         \rn{18}         & 69.68 & 70.38   & 69.76  & 68.32  & 70.43  & 70.01  & 68.83 \\
        \rn{50}         & 76.07 & 76.21   & 75.89   & 75.10   & 76.58   & 76.52   & 75.53      \\
        InceptionV3	        & 77.40 & 78.33 & 77.84 & 77.49 & 78.45 & 78.12 & 77.74 \\
        \mnv{2}      & 71.72 & 71.76   & 70.17  & 66.43  & 71.82   & 70.48  & 66.89  \\ 
        EfficientNet lite   & 75.42 & 75.17 & 71.55 & 70.22 & 74.75 & 73.92 & 71.55 \\
        DeeplabV3	        & 72.94 & 73.99 & 70.90 & 66.78 & 72.87 & 73.01 & 68.90 \\
        EfficientDet-D1	 & 40.08   & 38.94    & 35.34 & 24.70    & 38.97   &  36.75   &  28.68 \\
        BERT-base &  83.06  & 83.26  & 82.64  & 78.83 & 82.44  & 82.39  & 77.63\\
        \bottomrule
    \end{tabular}
     \caption{Performance (average over 3 runs) of our standard QAT pipeline for various models and tasks. DeeplabV3 (\mnv{2} backbone) is evaluated on Pascal VOC (mean intersection over union), EfficientDet-D1 on COCO 2017 (mean average precision), BERT-base on the GLUE benchmark and all other models on ImageNet (accuracy). We evaluate all models on the respective validation sets. Higher is better in all cases.}
     \label{tbl:standard_qat}
\end{table*}


\section{Summary and Conclusions}
\label{sec:conclusions}
Deep learning has become an integral part of many machine learning applications and can now be found in countless electronic devices and services, from smartphones and home appliances to drones, robots and self-driving cars. As the popularity and reach of deep learning in our everyday life increases, so does the need for fast and power-efficient neural network inference. Neural network quantization is one of the most effective ways of reducing the energy and latency requirements of neural networks during inference.

Quantization allows us to move from floating-point representations to a fixed-point format and, in combination with dedicated hardware utilizing efficient fixed-point operations, has the potential to achieve significant power gains and accelerate inference. However, to exploit these savings, we require robust quantization methods that can maintain high accuracy, while reducing the bit-width of weights and activations. 
To this end, we consider two main classes of quantization algorithms: Post-Training Quantization (PTQ) and Quantization-Aware Training (QAT).

Post-training quantization techniques take a pre-trained FP32 networks and convert it into a fixed-point network without the need for the original training pipeline. This makes them a lightweight, push-button approach to quantization with low engineering effort and computational cost. We describe a series of recent advances in PTQ and introduce a PTQ pipeline that leads to near floating-point accuracy results for a wide range of models and machine learning tasks. In particular, using the proposed pipeline we can achieve 8-bit quantization of weights and activations within only 1\% of the floating-point accuracy for all networks. We further show that many networks can be quantized even to 4-bit weights with only a small additional drop in performance. In addition, we introduce a debugging workflow to effectively identify and fix problems that might occur when quantizing new networks.

Quantization-aware training models the quantization noise during training through simulated quantization operations. This training procedure allows for better solutions to be found compared to PTQ while enabling more effective and aggressive activation quantization. Similar to PTQ, we introduce a standard training pipeline utilizing the latest algorithms in the field. We also pay special attention to batch normalization folding during QAT and show that simple static folding outperforms other more computationally expensive approaches. We demonstrate that with our QAT pipeline we can achieve 4-bit quantization of weights, and for some models even 4-bit activations, with only a small drop of accuracy compared to floating-point.

The choice between PTQ and QAT depends on the accuracy and power requirements of the application. Both approaches are an essential part of any model efficiency toolkit and we hope that our proposed  pipelines will help engineers deploy high-performing quantized models with less time and effort.

\bibliographystyle{icml2020}
\bibliography{references}

\cleardoublepage
\end{document}